%% file: VCM.tex
\pdfoutput=1

\documentclass[11pt]{article}

\usepackage[final]{acl}

\usepackage{xcolor}
\usepackage{times}
\usepackage{latexsym}
\usepackage{graphicx}
\usepackage{float}
\usepackage{colortbl}
\usepackage{multirow}
\usepackage{booktabs}
\usepackage{caption}
\usepackage{subcaption}
\usepackage{rotating}
\usepackage[export]{adjustbox}
\usepackage{dirtytalk} 
\usepackage[percent]{overpic}
\usepackage[normalem]{ulem}

\usepackage[table]{xcolor}
\usepackage{array}      
\usepackage{booktabs}   
\usepackage{graphicx}   
\usepackage[most]{tcolorbox}
\definecolor{D55E00}{HTML}{D55E00}
\definecolor{009E73}{HTML}{009E73}
\definecolor{0072B2}{HTML}{0072B2}

\usepackage{amsmath}
\usepackage{amssymb}

\usepackage{booktabs}
\usepackage[table]{xcolor}
\usepackage{siunitx}
\usepackage{caption}
\usepackage{graphicx}

\definecolor{Best}{HTML}{F2F7FD}

\definecolor{myboxcolor}{rgb}{0.402,0.402,0.402}
\newtcolorbox{mybox}[1][]{
        enhanced,
        title=#1,
        colback=myboxcolor!3,
        colbacktitle=myboxcolor!3,
        coltitle=black,
        left=4pt,
        right=4pt,
        top=4.5pt,
        bottom=0pt,
        attach boxed title to top left={xshift=8pt, yshift=-7pt},
        boxed title style={frame hidden, size=small, colback=myboxcolor!3},
        sharp corners,
        rounded corners,
        arc=7pt,
}

\usepackage{algorithm}
\usepackage{algorithmicx}   
\usepackage{algpseudocode}
\usepackage{float}
\usepackage{pdflscape}
\usepackage{hyperref}
\usepackage[]{acronym}
\usepackage{amsmath}

\usepackage{amssymb}

\usepackage{comment}
\usepackage[T1]{fontenc}

\usepackage[utf8]{inputenc}

\usepackage{microtype}

\usepackage{inconsolata}

\usepackage{graphicx}
\usepackage{enumitem}
\usepackage{makecell}

\title{Breaking the Likelihood Trap:\\ Variance-Calibrated Modulation for Large Language Model Decoding}

\author{
  Yuanhao Ding$^{1}$~~~
  Meimingwei Li$^{2}$~~~
  Esteban Garcés Arias$^{2,3}$~~~
  \textbf{Matthias Aßenmacher}$^{2,3}$\\~~~
  \textbf{Christian Heumann}$^{2}$~~~
  \textbf{Chongsheng Zhang}\thanks{\ \ Corresponding author}$^{1}$\\[1.5ex]
  $^1$School of Computer and Information Engineering, Henan University\\ $^2$Department of Statistics, LMU Munich,
  $^3$Munich Center for Machine Learning (MCML)\\[1.5ex]
  \url{{yhding, cszhang}@henu.edu.cn}, \quad \url{M.Li@campus.lmu.de}\\
  \url{{esteban.garcesarias, matthias, chris}@stat.uni-muenchen.de}
}

\begin{document}

\maketitle

\begin{abstract}
In open-ended generation, LLMs frequently fall into the ``\textit{likelihood trap}'', characterized by repetitive degeneration and vocabulary dullness, resulting in a discrepancy between machine-generated and human-written text. While post-hoc tail truncation (e.g., Top-$p$, Min-$p$) avoids sampling from the unreliable tail, it can misalign generation with human lexical preferences by over-sampling from the uncalibrated head; fixed scalar repetition penalties, in turn, ignore how the scale of the logit distribution varies across inference steps, which can disrupt semantic coherence. To address both shortcomings, we propose \textit{Variance-Calibrated Modulation} (VCM), a training-free pre-decoding intervention. VCM directly reshapes the probability distribution prior to truncation via two dynamic mechanisms: (1) \textit{Contextual Searchlight via PMI}, which naturally suppresses global stopwords and elevates context-evoked tokens, and (2) \textit{Adaptive Self-Debiasing}, which utilizes real-time logit standard deviation to provide scale-invariant penalization. In experiments across open-ended generation, factual QA, and mathematical reasoning, we show that VCM consistently mitigates the likelihood trap. With negligible computational overhead, VCM integrates with existing decoding strategies, improving diversity and coherence and, particularly at higher decoding temperatures, reasoning accuracy. Our code is publicly available on GitHub.\footnote{\url{https://github.com/AetherDing/VCM}}

\end{abstract}


\section{Introduction}
\label{sec:introduction}

\begin{figure}[!ht]
\centering
\includegraphics[width=.5\textwidth, trim = 0 0 0 .79cm, clip]{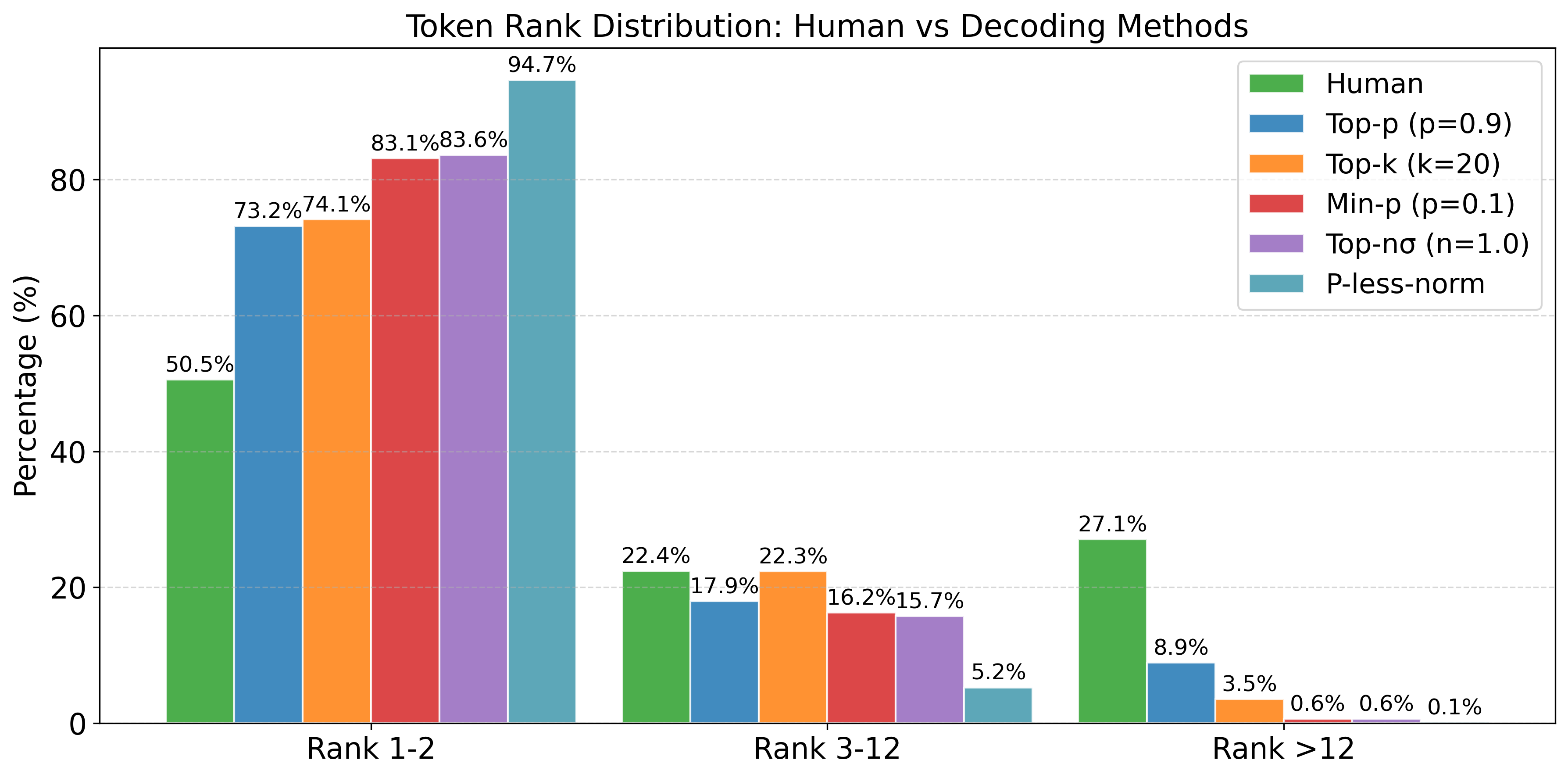}
\caption{\textbf{Token rank distribution of human-written text and standard decoding methods for Qwen3-8B.} Human next-token selections are distributed more broadly across the rank spectrum, with substantial mass on mid and lower-ranked candidates. In contrast, standard truncation-based strategies concentrate on the highest-ranked tokens, indicating a misalignment between humans and conventional decoding strategies.}
\label{fig:1}
\end{figure}

Despite the remarkable advancements of large language models (LLMs) in terms of zero-shot capabilities and for open-ended text generation, there remains a persistent discrepancy between machine-generated text and human-written text \cite{Reinhart2025,donmez-etal-2025-ai,zhou2026learntodistance}. When tasked with open-ended generation, LLMs frequently fall into the ``likelihood trap'' \cite{topp,chiang-chen-2021-relating,su2022contrastive,cd,zhu2024improving}, exhibiting two primary failure modes: \textit{degeneration}, i.e., getting stuck in mechanical, repetitive loops, and \textit{dullness}, i.e., over-relying on safe, generic, and high-frequency vocabulary \cite{li-etal-2022-evade,xu-etal-2023-look,zhu-etal-2023-penalty,acs2024,yao-etal-2025-understanding,ding-etal-2025-guard}.

\begin{figure*}[!ht]
\centering
\includegraphics[width=\linewidth, trim = 0 20cm 0 2cm, clip]{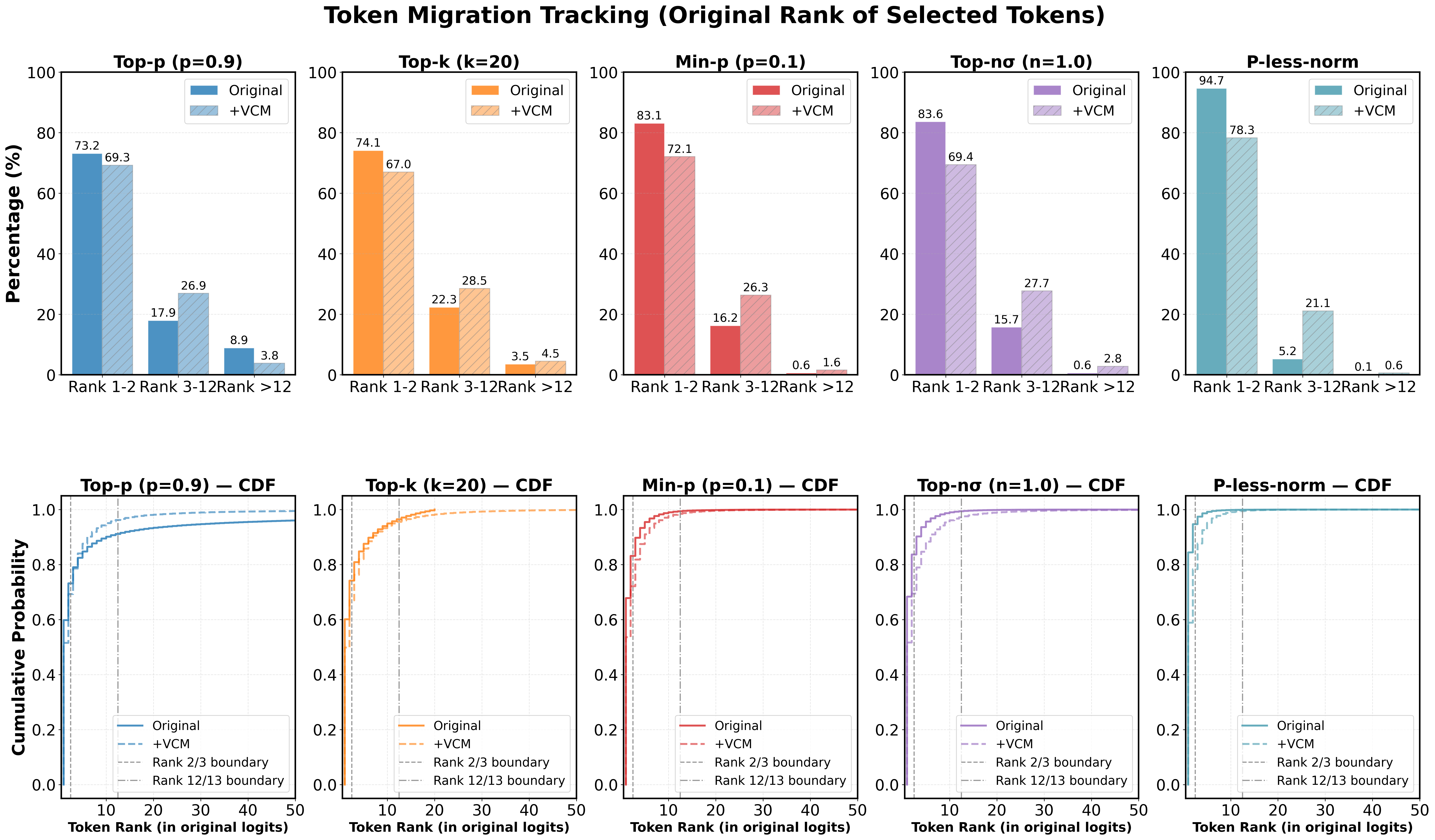}
\caption{\textbf{Token migration across decoding methods.} After applying VCM, token selections shift consistently from Ranks 1--2 toward 3--12 across strategies, producing a rank profile closer to the human distribution (cf. Fig. \ref{fig:1}).}
\label{fig:2}
\end{figure*}

To tackle these issues, modern generation pipelines heavily rely on stochastic decoding strategies, such as Top-$k$ \cite{topk}, Top-$p$ \citep[nucleus,][]{topp}, Min-$p$ \cite{minp}, Top-$n\sigma$ \cite{topn} or $p\textrm{-less}$ sampling \cite{pless}, all of which rely on \textit{post-hoc tail truncation}: They filter out the unreliable ``tail'' of the probability distribution before sampling. We empirically show that this truncation-only approach is misaligned with human (lexical) preferences using a 500-sample subset of the open-domain WikiText dataset \cite{wikitext1314} (about 100{,}000 token-level selection steps); the full 1{,}314-sample set is used for the main evaluation (\S\ref{sec:exp_setup}). As illustrated in Figure \ref{fig:1}, standard decoding methods exhibit an extreme bias towards the absolute top-ranked candidates (Rank 1-2). In contrast, human text generation is notably more diverse; humans frequently select tokens from the middle-tier (Rank 3-12) or even the tail (Rank $>12$), rather than strictly defaulting to high-probability tokens at every step; \citet{garces-arias-2026-blindspot} quantify this \emph{truncation blind spot} at scale, finding that 8--18\% of human-selected tokens fall outside common truncation boundaries across five open models and three domains. By merely truncating the tail, existing strategies force the model to continuously sample from an overly dense, uncalibrated distribution head, inadvertently exacerbating dullness and predictability. Simultaneously, existing pre-decoding interventions (e.g., repetition penalties) attempt to penalize previously generated tokens, but rely almost exclusively on fixed scalar subtractions. This approach ignores the fact that the scale of logits varies drastically across models and across individual inference steps, from highly confident/sharp distributions to highly uncertain/flat distributions. A fixed penalty applied to a flat (high entropy) distribution can severely disrupt semantic coherence and induce hallucinations, whereas the same penalty on a sharp (low entropy) distribution often proves insufficient to break repetitive loops.

Motivated by this, we propose \textbf{VCM (Variance-Calibrated Modulation)}, a novel, training-free, pre-decoding, logit-level intervention framework. VCM directly reshapes the probability distribution before any truncation is applied, simultaneously addressing both dullness and repetition through two dynamic mechanisms: (1) \textit{Contextual Searchlight:} By subtracting a pre-computed prior distribution (e.g., the unconditional logits from an empty context) from the current logits, VCM naturally suppresses globally high-frequency ``stopwords'' and dynamically elevates tokens that are uniquely evoked by the current context. (2) \textit{Adaptive Self-Debiasing:} To solve the scale-variance problem, we utilize the real-time standard deviation of the logits ($\sigma_t$) as the base unit for penalization. This ensures the intervention is scale-invariant, proportionally scaling up the penalty when the model exhibits high confidence (to break loops) and relaxing it when the model is uncertain. Figure \ref{fig:2} shows that integrating VCM with existing decoding methods successfully mitigates the over-concentration at the head of the distribution. It systematically redistributes the probability mass from the absolute top (Rank 1-2) to the middle tiers (Rank 3-12), resulting in a rank distribution that closely mimics the human preference curve observed in Figure \ref{fig:1}. We investigate one apparent artifact---that combining VCM with Top-$p$ sampling paradoxically reduces selections in the extreme tail (Rank $>12$)---in Appendix~\ref{app:vcm and topp}. We find that this is not a compatibility failure but a direct consequence of VCM concentrating probability mass on high-quality, context-relevant tokens, which causes the Top-$p$ threshold to be reached notably faster.

\paragraph{Contributions.} 

\begin{itemize}
    \item We propose \textit{Variance-Calibrated Modulation}, a training-free method that intervenes directly in the logit space prior to truncation, mitigating the ``likelihood trap'' and reducing front-heavy sampling biases (\S\ref{sec:3.1}).
    \item We use pointwise mutual information (PMI) to construct a \textit{Contextual Searchlight} elevating context-evoked tokens (\S \ref{sec:3.2}), alongside a scale-invariant \textit{Adaptive Self-Debiasing} penalty to disrupt loops without destroying coherent logic (\S \ref{sec:3.3}). We also uncover its implicit temperature-scaling effect (\S \ref{sec:3.4}).
    \item VCM improves the balance across open-ended generation, factual QA, and mathematical reasoning (\S \ref{sec:5.1}--\ref{sec:5.3}) with negligible latency overhead ($\sim\!1.01\times$) compared to existing baselines (\S \ref{sec:5.4}). Additional analysis indicates that VCM shifts generation toward human lexical distributions (Appendix~\ref{app:tokenrankcdf}).
\end{itemize}


\section{Related Work}
\label{sec:related}

We situate VCM within the context of two major streams of existing decoding strategies: probability-based truncation sampling and logit-space interventions with dynamic penalties.

\paragraph{Probability-based Truncation Sampling.} LLMs typically generate text by sampling from the probability distribution over the vocabulary. Classical approaches seek to balance diversity and determinacy by reshaping or constraining this distribution. For instance, Top-$k$ sampling \cite{topk} restricts the candidate set to the $k$ most probable tokens, whereas Top-$p$ (nucleus) sampling \cite{topp} adaptively determines the candidate set via a cumulative probability threshold. Locally typical sampling \cite{typical} instead selects 
tokens whose surprisal is close to the conditional entropy of the distribution. To mitigate the instability induced by high-temperature sampling, a range of adaptive truncation schemes has recently emerged. Min-$p$ \cite{minp} establishes a dynamic lower bound relative to the most probable token; Top-$n\sigma$ \cite{topn} decouples the effect of temperature using the maximum logit and the global standard deviation; and Min-$k$ \cite{ding2026mink} as well as $p\textrm{-less}$ sampling \cite{pless} dynamically locate the ``semantic cliff'' or define hyperparameter-free truncation thresholds by leveraging local relative decay and collision entropy (i.e., second-order information), respectively. Although these recent methods have achieved notable progress in robustness and inference efficiency, they all share a common limiting paradigm: \emph{post-hoc tail truncation}. As noted in Section \ref{sec:introduction}, these methods merely discard the unreliable ``long tail'' after applying the softmax, leaving a measurable share of human-selected tokens unreachable \citep{garces-arias-2026-blindspot}, while failing to address the more fundamental issue that probability mass is often excessively concentrated in the absolute head of the distribution. This truncation frequently leads the model into a ``likelihood trap'', substantially exacerbating dullness and mediocrity in generated text by overlooking relevant middle-tier tokens.

\paragraph{Logit-space Interventions and Dynamic Penalties.} To reduce the model's over-reliance on generic high-frequency tokens, prior work has introduced distributional contrasts during decoding. Contrastive Decoding \citep[CD,][]{cd} and Classifier-Free Guidance \cite{sanchez2024stay} amplify context-specific knowledge by penalizing tokens preferred by an ``amateur'' or unconditional model. PMI has also been theoretically motivated as a mechanism for promoting context-specific tokens \cite{nandwani-etal-2023-pointwise}. However, such methods typically require either additional forward passes or intricate heuristic rescaling, resulting in substantial computational overhead. In contrast, our VCM method directly subtracts a precomputed unconditional prior in the original logit space. Owing to Softmax's translation invariance, this operation serves as an exact proxy for contextual PMI, thereby natively neutralizing globally frequent stopwords without relying on any external model.

On the other hand, to avoid degeneration, conventional decoding strategies commonly rely on fixed scalar penalties, such as the repetition penalty \cite{keskar2019,topp,zhu-etal-2023-penalty}, yet they ignore the model's scale variance across decoding steps, and may easily disrupt semantic coherence when the distribution becomes flat. To introduce dynamic adaptability, uncertainty-aware strategies---including Mirostat \cite{mirostat}, $\eta$-sampling \cite{eta-sampling}, $\epsilon$-sampling \cite{epsilon}, and REAL \cite{real}---adjust truncation boundaries or temperature on the fly. Moreover, advanced decoding strategies such as ACS \cite{acs2024}, GUARD \cite{ding-etal-2025-guard}, and G2 \cite{g2} employ token-level entropy as a global gating signal to regulate contrastive search \citep[CS,][]{su2022contrastive} parameters or diversity prompts. While theoretically appealing, entropy-based methods often introduce additional computational cost and complex hyperparameter tuning \cite{zhou-etal-2025-balancing,minp,topn,ding2026mink,li2026}. By directly using the real-time standard deviation of the logits, $\sigma_t$, VCM achieves scale invariance. It strongly amplifies penalties during high-confidence repetitive loops, while relaxing them smoothly when the model is uncertain, thus offering a robust, training-free alternative to both rigid scalar penalties and unstable entropy-based computations.

\paragraph{Adaptive Pre-Decoding.} Recent methods have begun addressing the factuality-diversity trade-off through inference-time control. For instance, Dynamic Focus Decoding \cite[DFD,][]{luo-etal-2025-odysseus} dynamically adjusts step-level temperature based on layer-wise hidden-state dynamics. While both DFD and VCM are training-free, they intervene at fundamentally different levels. VCM provides a computationally lighter intervention at the token-level in logit space prior to Softmax and truncation. By leveraging contextual priors and variance-calibrated repetition modulation, VCM bypasses the need for deep hidden-state extraction, offering a highly efficient alternative to adaptive focus control.


\section{Methodology: The VCM Framework}
\label{sec:methodology}

In this section, we formally introduce \textbf{VCM}, our training-free logits modulation framework.

\subsection{Preliminaries and Problem Formulation}
\label{sec:3.1}

Given a sequence of context tokens $x_{<t} = (x_1, \dots, x_{t-1})$, an autoregressive language model $\mathcal{M}$ computes a dense logits vector $z_t \in \mathbb{R}^{|\mathcal{V}|}$ over the vocabulary $\mathcal{V}$. The next-token probability distribution is obtained via the Softmax function:
\begin{equation}
P(x_t = w_i \mid x_{<t}) =
\frac{\exp(z_{t,i})}{\sum_{j=1}^{|\mathcal{V}|} \exp(z_{t,j})},
\quad \forall w_i \in \mathcal{V}.
\end{equation}

Decoding interventions relying on repetition penalties traditionally apply a fixed scalar subtraction to the logits of previously generated tokens. However, this fixed-penalty paradigm ignores the fact that the absolute magnitude and variance of $z_t$ vary drastically with the model's confidence at step $t$. A fixed penalty might be negligible when the distribution is sharp, yet it catastrophically disrupts semantic coherence when the distribution is flat. VCM addresses this by introducing an adaptive, context-aware modulation signal $\mathcal{I}_t$:
\begin{equation}
\tilde{z}_t = z_t + \alpha \cdot \mathcal{I}_t
\end{equation}
where $\tilde{z}_t$ represents the modified logits, and $\alpha \in [0,1]$ is a hyperparameter controlling the intervention strength. In practice, the preferred choice of $\alpha$ is task-dependent rather than universal. Therefore, instead of relying on a single global setting, we identify stable task-appropriate operating regions through systematic sensitivity analysis; details are provided in Appendix~\ref{app:alpha choice}.

The signal $\mathcal{I}_t$ comprises two complementary components: a contextual PMI term ($\mathcal{S}_{\text{prior}}$, \S\ref{sec:3.2}) and an adaptive self-debiasing term ($\mathcal{C}_{\text{rep}}$, \S\ref{sec:3.3}). Our ablation study (Appendix~\ref{sec:ablation_components}) indicates that the two terms are complementary and that the full model is needed to balance diversity against factual and reasoning accuracy.

\subsection{Contextual Searchlight via PMI}
\label{sec:3.2}

To mitigate generation dullness and the over-reliance on generic high-frequency words (e.g., ``the'', ``is''), we aim to promote tokens that are highly specific to the current context $x_{<t}$. We formalize this using an approximation of PMI, which measures the strength of association between the context and a token:
\begin{multline}
\text{PMI}(w_i; x_{<t}) =
\log \frac{P(w_i \mid x_{<t})}{P(w_i)} \\
\propto
\log P(w_i \mid x_{<t}) - \log P(w_i).
\end{multline}

Since an LLM's raw outputs serve as unnormalized log-probabilities, we can approximate the context-conditioned term $\log P(w_i \mid x_{<t})$ using the current logits $z_t$. To approximate the marginal prior $\log P(w_i)$, we pre-compute a prior logits vector $z_{\text{prior}}$ by executing an unconditional forward pass (e.g., using an empty string or a solitary BOS token). We systematically investigate the choice of this unconditional prior, finding that no single prior dominates across tasks; we retain the BOS token as a semantically minimal, task-agnostic default (cf. Appendix \ref{sec:ablation_prior}).

We define the contextual searchlight score $\mathcal{S}_{\text{prior}} \in \mathbb{R}^{|\mathcal{V}|}$ in the unnormalized logit space:
\begin{equation}
\mathcal{S}_{\text{prior}} = z_t - z_{\text{prior}}.
\end{equation}

While $z_t$ and $z_{\text{prior}}$ contain distinct partition functions (normalization constants), subtracting them introduces a uniform constant shift across the vocabulary. Crucially, because the standard Softmax operation is translation-invariant, this constant shift is mathematically canceled out during the final probability conversion. Thus, $\mathcal{S}_{\text{prior}}$ serves as an exact, stable proxy for the context-conditioned PMI, elevating tokens uniquely evoked by $x_{<t}$ while naturally neutralizing globally high-frequency generic tokens.

\subsection{Adaptive Self-Debiasing via Distributional Variance}
\label{sec:3.3}

To address the degeneration problem while avoiding fixed-scalar penalties, we introduce a self-scaling penalization mechanism based on the real-time statistical properties of the logits vector.

At time step $t$, we compute the standard deviation of the current logits vector $z_t$:
\begin{equation}
\sigma_t =
\sqrt{
\frac{1}{|\mathcal{V}|}
\sum_{i=1}^{|\mathcal{V}|}
(z_{t,i} - \bar{z}_t)^2
}
\end{equation}

We utilize $\sigma_t$ as the base unit of magnitude for penalization. Let $W$ be a predefined local context window size\footnote{In all experiments, we set the context window size $W = 128$, which we empirically find to be robust across tasks (see Appendix \ref{sec:ablation_window} for a more detailed analysis).}. We define the frequency of a token $w_i$ in the recent context as
\[
c_i = \mathrm{Count}(w_i, x_{t-W:t-1}).
\]

The adaptive repetition cost vector $\mathcal{C}_{\text{rep}} \in \mathbb{R}^{|\mathcal{V}|}$ is defined element-wise as
\begin{equation}
\mathcal{C}_{\text{rep}, i} = c_i \cdot \sigma_t.
\end{equation}

\paragraph{Theoretical Advantage (Scale Invariance).}
This formulation guarantees that the penalty remains proportional to the model's current confidence. When the model is certain (large $\sigma_t$), the penalty scales up proportionately to forcefully break a repetition loop. Conversely, when the model is uncertain (small $\sigma_t$), the penalty decays smoothly, safeguarding fragile, low-confidence token predictions from undeserved suppression.

\subsection{Unified Modulation and Decoding}\label{sec:3.4}

By combining contextual prior and calibrated repetition cost, we define the VCM intervention as:
\[
\mathcal{I}_t = \mathcal{S}_{\text{prior}} - \mathcal{C}_{\text{rep}}.
\]
The reshaped logits are thus computed as:
\begin{equation}
\tilde{z}_t
=
z_t + \alpha \cdot \bigl[(z_t - z_{\text{prior}}) - \mathcal{C}_{\text{rep}}\bigr].
\end{equation}

Expanding this formulation reveals a useful mechanistic side-effect:
\begin{equation}
\tilde{z}_t
=
(1+\alpha)z_t
- \alpha \cdot z_{\text{prior}}
- \alpha \cdot \mathcal{C}_{\text{rep}}.
\end{equation}

The coefficient $(1+\alpha)$ applied to $z_t$ acts as an \textbf{implicit temperature scaling} with
\[
T = \frac{1}{1+\alpha}.
\]

This implicit sharpening helps explain the empirical interaction between VCM and nucleus sampling documented in Appendix~\ref{app:vcm and topp}. As VCM extracts the high-quality signal, it simultaneously sharpens the probability landscape, forcing the Top-$p$ probability mass to reach its threshold earlier, naturally truncating the noisy tail without requiring manual temperature tuning. The modulated logits $\tilde{z}_t$ are subsequently passed through the Softmax function to obtain the reshaped probability distribution, which can be seamlessly plugged into any standard post-hoc truncation method (e.g., Top-$p$, Top-$k$). The full procedure is summarized in Algorithm~1.

\begin{algorithm}[t]
\caption{VCM Decoding}
\begin{algorithmic}[1]
\Require Model $\mathcal{M}$, context $x_{<t}$, prior $z_{\text{prior}}$, mixing coefficient $\alpha$, window size $W$
\Ensure Next token $x_t$
\State $z_t \gets \mathcal{M}(x_{<t})$
\State $\mathcal{S}_{\text{prior}} \gets z_t - z_{\text{prior}}$
\State $\sigma_t \gets \mathrm{StandardDeviation}(z_t)$
\State $\mathcal{C}_{\text{rep}} \gets \vec{0} \in \mathbb{R}^{|\mathcal{V}|}$
\For{each token $w$ in $x_{t-W:t-1}$}
    \State $\mathcal{C}_{\text{rep}}[w] \gets \mathrm{Count}(w, x_{t-W:t-1}) \cdot \sigma_t$
\EndFor
\State $\mathcal{I}_t \gets \mathcal{S}_{\text{prior}} - \mathcal{C}_{\text{rep}}$
\State $\tilde{z}_t \gets z_t + \alpha \cdot \mathcal{I}_t$
\State \Return $\mathrm{Sampler}(\mathrm{softmax}(\tilde{z}_t))$
\end{algorithmic}
\end{algorithm}


\section{Experimental Setup}
\label{sec:exp_setup}

\paragraph{Models.} To assess our method and its applicability across models, we conduct experiments on four open-source models from two model families: Qwen3-8B\footnote{https://huggingface.co/Qwen/Qwen3-8B} and Qwen3-4B-Instruct\footnote{https://huggingface.co/Qwen/Qwen3-4B-Instruct-2507} \cite{qwen3}, as well as LLaMA-3-8B\footnote{https://huggingface.co/meta-llama/Meta-Llama-3-8B} and LLaMA-3-8B-Instruct\footnote{https://huggingface.co/meta-llama/Meta-Llama-3-8B-Instruct} \cite{llama3}. The main experiments use Qwen3-8B for WikiText open-ended generation and LLaMA-3-8B-Instruct for TruthfulQA, TriviaQA, GSM8K, and MATH500; LLaMA-3-8B and Qwen3-4B-Instruct are used in the cross-model experiments (\S\ref{sec:cross_model}).

\paragraph{Datasets.} To evaluate the generalizability of our method across different tasks, we use benchmark datasets covering three representative categories of downstream tasks. Specifically, for the open-ended text generation task, we use WikiText \citep[1{,}314 samples,][]{wikitext1314}, for question answering (QA), we use TruthfulQA \citep[817 samples,][]{truthfulqa} and TriviaQA \citep[1{,}500 samples,][]{triviaqa}, and for (mathematical) reasoning, we use GSM8K \citep[1{,}319 samples,][]{gsm8k} and MATH500 \citep[500 samples,][]{math500,math}. For the reasoning tasks, we preserve the complete reasoning trace and automatically extract the final answer for evaluation, following \citet{topn}, to assess both the quality of the reasoning process and the correctness of the final prediction.

\paragraph{Baselines.} Our comparison focuses on sampling-based decoding strategies and lightweight pre-decoding interventions, as VCM is designed as a plug-in logit-space module for standard stochastic samplers. We compare it with a set of widely used sampling methods: Top-$k$ ($k=20$), Top-$p$ ($p=0.9$), $\eta$-sampling ($\eta=3 \times 10^{-4}$), $\epsilon$-sampling ($\epsilon=9 \times 10^{-4}$), locally typical sampling ($\tau=0.9$), Min-$p$ ($p=0.1$), Top-$n\sigma$ ($n=1.0$), Min-$k$ ($\tau=3.0$), $p\textrm{-less}$ sampling, CD ($\alpha$ = 0.5, $\beta$ = 0.7, amateur-T = 1.0, amateur = Qwen3-0.6B (WikiText) / LLaMA-3.2-1B-Instruct (QA and math tasks)), DFD ($\alpha=0.1$, $\sigma=0.4$, exponential). For all baseline samplers, we use the recommended configurations from prior empirical studies and official implementations \cite{minp,topn,garces-arias-etal-2025-decoding,luo-etal-2025-odysseus,ding2026mink}. 

\paragraph{On the choice of $\alpha$.} When VCM is combined with a sampler, the sampler-specific hyperparameters are kept unchanged; VCM only introduces the mixing coefficient $\alpha$. This coefficient $\alpha$ governs the strength of VCM and mediates the trade-off between conservative decoding and aggressive distribution reshaping. Empirically, we find that VCM is not overly sensitive to a single exact value but instead performs well within task-specific intervals, so rather than tuning $\alpha$ to a single brittle optimum, we identify stable task-level operating ranges and select representative values within these ranges for the main experiments. This behavior suggests that $\alpha$ should be understood as a controllable operating regime rather than a dataset-agnostic constant. Details on the choice of $\alpha$ when combining VCM with different methods across various tasks are provided in Appendix \ref{app:alpha choice}.

\paragraph{Evaluation Metrics.} We use different evaluation metrics for different task types. For open-ended text generation, we report Rep-$n$ \cite{Welleck2020Neural}, Distinct-$n$ \cite{li-etal-2016-diversity}, MAUVE \cite{mauve}, BERTScore \cite{BERTScore:}, LLM-as-a-Judge (DeepSeek V3.2~\citep{deepseekv32} and Qwen3.6-Plus; prompt and Cohen's $\kappa$ in Appendix~\ref{app:llmjudge}). For QA, TruthfulQA is evaluated in terms of factual accuracy (Truthfulness \& Informativeness) and Type-Token Ratio (TTR), while TriviaQA is evaluated using Exact Match (EM) and F1. For reasoning tasks, we use EM as a metric.


\begin{table*}[!htb]
\centering
\resizebox{1\textwidth}{!}{\input{fig_tables/table1_wikitext}}
\caption{Open-ended generation on WikiText (Qwen3-8B). Means are shown with $\pm$ $95\%$ confidence intervals; \textbf{bold} marks the better value in each baseline/+VCM pair. Rep-2 and BERTScore (per-sample, $N{=}1314$) use a paired $t$-test, and the LLM-as-a-Judge win rates a binomial sign test over non-tied comparisons (${}^{*}/{}^{**}/{}^{***}$: $p<0.05/0.01/0.001$). MAUVE and Distinct-2 are aggregate summaries reported without per-sample intervals; the MAUVE gain is consistent across all nine samplers (paired Wilcoxon, $p=0.004$).}
\label{tab:table1}
\end{table*}

\begin{table}[ht]
\centering
\resizebox{0.49\textwidth}{!}{\input{fig_tables/table2_QA}}
\caption{Question answering with LLaMA-3-8B-Instruct. Means with $\pm$ $95\%$ confidence intervals; \textbf{bold} marks the better value per pair. Significance of the +VCM gain: paired $t$-test (TTR), McNemar's test (Factuality), two-proportion test (EM) (${}^{*}/{}^{**}/{}^{***}$: $p<0.05/0.01/0.001$). TTR improves significantly for every sampler; Factuality and EM gains are positive for all nine samplers, reach per-sampler significance only for $\eta$-sampling, and show a significant overall direction ($p<0.01$).}
\label{tab:table2}
\end{table}

\section{Results}
\label{sec:results}

To rigorously evaluate the proposed VCM framework, we conduct experiments across three task families: open-ended generation, factual QA, and mathematical reasoning. Alongside the quantitative results presented below, we provide comprehensive qualitative case studies (Appendix \ref{app:case}, Tables~\ref{tab:triviaqa case}--\ref{tab:gsm8k case}), which illustrate how VCM reduces degenerate loops, supports factual precision, and preserves step-by-step coherence in representative examples.

\subsection{Open-ended Text Generation}
\label{sec:5.1}

As shown in Table~\ref{tab:table1}, the standard truncation baselines exhibit clear signs of the ``likelihood trap'', producing repetitive and lexically narrow text (high Rep-2, low Distinct-2, and MAUVE). Integrating VCM yields a consistent shift across all decoding backbones: it substantially reduces repetition (Rep-2) and increases lexical diversity (Distinct-2). The accompanying gains in BERTScore are statistically significant for every sampler (paired $t$-test, $p<10^{-90}$; Cohen's $d_z=0.66$--$0.81$), and the large, consistent rise in MAUVE (mean $2.21\rightarrow14.04$ across the nine samplers; paired Wilcoxon $p=0.004$) indicates that this added diversity does not come at the expense of coherence, but instead reflects more contextually relevant vocabulary. Both LLM judges also prefer VCM outputs over their baselines in every setting (binomial sign test on non-tied comparisons, $p<0.05$). Together, these results suggest that VCM shifts the generation distribution toward that of human text, an alignment further illustrated by the token-rank CDF analysis in Appendix~\ref{app:tokenrankcdf}.

\subsection{Question Answering}
\label{sec:5.2}

We next consider factual QA (Table~\ref{tab:table2}). On TruthfulQA, VCM consistently improves lexical diversity, yielding statistically significant gains in TTR across all samplers. Improvements in Factuality (TruthfulQA) and Exact Match (TriviaQA) are positive for all nine samplers; they reach per-sampler significance only for $\eta$-sampling (Factuality, McNemar's test, $p=0.017$; EM, two-proportion test, $p=0.012$), while for the remaining samplers they remain within the corresponding $95\%$ confidence intervals. Still, the uniform positive direction across all nine samplers (sign test, $p<0.01$) suggests a moderate but systematic benefit. Overall, the results indicate that VCM enhances diversity without sacrificing factual accuracy, likely because the contextual prior $\mathcal{S}_{\text{prior}}$ promotes query-specific tokens over generic fallback continuations.

\begin{table}[t]
\centering
\resizebox{0.49\textwidth}{!}{\input{fig_tables/table3_reasoning}}
\caption{Exact-Match accuracy on GSM8K and MATH500 (LLaMA-3-8B-Instruct) at standard ($T{=}1.0$) and high ($T{=}2.0$) temperature. Means with $\pm$ $95\%$ confidence intervals; ${}^{*}/{}^{**}/{}^{***}$ mark a significant +VCM gain (two-proportion test, $p<0.05/0.01/0.001$). Gains are positive in all settings ($p<0.01$) and are largest, and most often significant, at $T{=}2.0$.}
\label{tab:table3}
\end{table}

\begin{table}[!htb]
\centering
\resizebox{0.49\textwidth}{!}{\input{fig_tables/table4_com_DFD}}
\caption{VCM compared with Contrastive Decoding (CD) and Dynamic Focus Decoding (DFD), all applied over a common Top-$p$ backbone. EM columns show means with $\pm$ $95\%$ confidence intervals; ${}^{*}/{}^{**}/{}^{***}$ mark a significant gain over the block's baseline sampler (two-proportion test). MAUVE and F1 are reported without intervals. \textbf{Bold} marks the best value per column within each block.}
\label{tab:table4}
\end{table}

\subsection{Mathematical Reasoning}
\label{sec:5.3}

Tasks that demand step-by-step logic, such as GSM8K and MATH500, are sensitive to logit interventions, since over-penalization can corrupt a reasoning chain. As shown in Table~\ref{tab:table3}, VCM matches or improves accuracy in every setting, with EM positive across all nine samplers (cross-sampler sign test, $p<0.01$). The improvements are small at standard temperature ($T{=}1.0$) and largest---reaching per-cell significance for several samplers---at high temperature ($T{=}2.0$), where baseline decoding degrades most. This pattern is consistent with the scale-invariant design of VCM: with a conservative mixing coefficient ($\alpha\le0.2$ in all GSM8K settings and at $T{=}1.0$ on MATH500), the intervention is intended to nudge the model out of degenerate loops while leaving the repeated symbols and operands required for valid derivations largely intact.

\subsection{Comparison with other Pre-Decoding Interventions}
\label{sec:5.4}

\paragraph{Performance.}

Table~\ref{tab:table4} compares VCM with two recent pre-decoding interventions, Contrastive Decoding (CD) and Dynamic Focus Decoding (DFD), all applied over a common Top-$p$ backbone. CD and DFD improve over the baseline mainly on the QA and reasoning metrics but yield only modest gains in open-ended diversity. VCM shows the opposite, and complementary, profile: it produces by far the largest MAUVE gains on WikiText, while remaining competitive with CD and DFD on TriviaQA EM/F1 and GSM8K EM, where the differences among the three interventions are small and mostly within confidence intervals. Overall, VCM offers a favorable balance across the three task types; its computational cost, analyzed next, is minimal.

\paragraph{Speed and Efficiency.}
For fair comparison, all methods use Top-$p$ ($p=0.9$) as the unified sampling backbone. CD, DFD, and VCM each intervene at the logit level prior to Top-$p$ truncation and sampling. A critical bottleneck of existing intervention strategies is inference latency. As reported in Table \ref{tab:decoding-speed}, CD requires an auxiliary forward pass, incurring a roughly $1.5\times$ latency penalty, and DFD relies on intensive layer-wise hidden-state dynamics, causing over a $2.5\times$ slowdown. In stark contrast, VCM operates via a closed-form, element-wise logit manipulation prior to Softmax. Consequently, VCM attains its gains with negligible computational overhead ($\sim\!1.01\times$ the baseline), essentially preserving standard auto-regressive decoding speed.

Generation-length statistics are reported in Appendix~\ref{app:generation_length}. VCM produces shorter outputs on TriviaQA and only moderately longer outputs on GSM8K, whereas WikiText uses a fixed generation budget, suggesting that the observed gains are not attributable to systematically longer generations.

\subsection{Cross-Model Transfer}
\label{sec:cross_model}

The main experiments use Qwen3-8B for WikiText and LLaMA-3-8B-Instruct for the QA and reasoning benchmarks. To further assess whether the effectiveness of VCM generalizes beyond these primary model settings, we conduct additional experiments with two different models: LLaMA-3-8B on WikiText and Qwen3-4B-Instruct on GSM8K. These experiments are intended as additional cross-model evaluations rather than a reversal of the task--model assignments used in the main experiments. We consider Top-$p$, Top-$k$, and Min-$p$, using the same baseline versus $+\mathrm{VCM}$ comparison as in the main experiments.

As shown in Table~\ref{tab:cross_model_transfer}, VCM improves all reported metrics for all three samplers across both additional model settings. On WikiText, VCM consistently increases MAUVE and BERTScore while reducing Rep-2 with LLaMA-3-8B. On GSM8K, it improves exact-match accuracy with Qwen3-4B-Instruct under all three samplers. These results provide further evidence that the benefits of VCM are not limited to the specific models used in the main experiments and can generalize across different model families, sizes, and model variants.

\begin{table}[t]
\centering
\resizebox{0.49\textwidth}{!}{
\input{fig_tables/table5_speed}
}
\caption{We report the average latency per generated token (ms/tok) and the overhead ratio relative to the standard Top-$p$ baseline. The maximum generation length is set to 256, 512, and 1024 for WikiText, TriviaQA, and GSM8K, respectively.}
\label{tab:decoding-speed}
\end{table}

\begin{table}[t]
\centering
\resizebox{.49\textwidth}{!}{
\input{fig_tables/table6_crossmodel}
}
\caption{Cross-model transfer results with additional models. WikiText is evaluated with LLaMA-3-8B ($\alpha=0.1$ and temperature =1.0), while GSM8K is evaluated with Qwen3-4B-Instruct ($\alpha=0.1$ and temperature =1.0). VCM improves every reported metric across Top-$p$, Top-$k$, and Min-$p$.}
\label{tab:cross_model_transfer}
\end{table}


\section{Conclusion}

We have examined two recurring weaknesses of modern decoding---the head-over-concentration induced by \textit{post hoc} tail truncation and rigid repetition penalties---both of which push LLMs into mechanical, repetitive loops. To address them, we introduce Variance-Calibrated Modulation (VCM), a lightweight, training-free pre-decoding intervention. By pairing a PMI-based contextual prior with a variance-scaled repetition penalty, VCM reshapes the logit distribution to counter the likelihood trap before any truncation is applied. Across open-ended generation, factual QA, and mathematical reasoning, our results indicate that VCM moves machine output closer to human lexical distributions, improving diversity and coherence (with significant gains in BERTScore, MAUVE, and TTR) while preserving---and at higher temperatures improving---reasoning accuracy. At a latency cost of only $\sim\!1.01\times$, VCM offers an efficient, plug-and-play addition to existing samplers.


\section*{Limitations}

Despite its simplicity and effectiveness, VCM has several limitations:

First, our experiments are conducted on medium-scale open-source models, with WikiText evaluated on Qwen3-8B and LLaMA-3-8B, and the QA and reasoning tasks evaluated on LLaMA-3-8B-Instruct and Qwen3-4B-Instruct. Further evaluation is needed on larger, smaller, multilingual, and domain-specialized models.

Second, VCM introduces a task-dependent mixing coefficient $\alpha$. Our sensitivity analysis shows that performance is stable within broad operating ranges, but the preferred range differs across tasks: open-ended generation benefits from stronger modulation, factual QA requires moderate intervention, and mathematical reasoning generally needs conservative settings. Automatically selecting or scheduling $\alpha$ remains an important direction for future work.

Third, improved diversity does not necessarily imply improved factuality. While VCM reduces repetition and often improves QA and reasoning accuracy, open-ended generation may still produce fluent but unsupported continuations. For hallucination-sensitive applications, VCM should be combined with factuality-aware evaluation, retrieval grounding, or verification mechanisms.

Finally, our main comparisons focus on sampling-based decoding and lightweight pre-decoding interventions. We do not exhaustively compare with methods based on reranking, additional training, prompt-level control, or task-specific supervision, which operate under different computational and supervision assumptions. Moreover, although we use automatic metrics and LLM-as-a-Judge evaluations, more extensive human evaluation, as well as multicriteria evaluation frameworks that account for trade-offs among metrics \citep{garces-arias-etal-2025-statistical}, would further strengthen the assessment of coherence, factuality, and user-perceived quality.

\section*{Ethics Statement}

We affirm that our research adheres to the \href{https://www.aclweb.org/portal/content/acl-code-ethics}{ACL Ethics Policy}. This work involves the use of publicly available datasets and does not include any personally identifiable information. An ethical concern worth mentioning is the use of language models for text generation, which may produce harmful content, either through intentional misuse by users or unintentionally due to the training data or algorithms. We declare that there are no conflicts of interest that could potentially influence the outcomes, interpretations, or conclusions of this research. All funding sources supporting this study are enumerated in the acknowledgments section. We have diligently documented our methodology, experiments, and results, and we commit to sharing our code, data, and other relevant resources to enhance reproducibility and further advancements in the field.

\section*{Acknowledgments}
This work was partially supported by the National Natural Science Foundation of China (62676141, 62250410371, W26332188), the MOE Liberal Arts and Social Sciences Foundation (No.23YJAZH210), the Major Program of National Social Science Foundation (No.23\&ZD309), and Henan Provincial Center for Outstanding Overseas Scientists (No.GZS2025004). Matthias A\ss{}enmacher received funding from the Deutsche Forschungsgemeinschaft (DFG, German Research Foundation) under the National Research Data Infrastructure-NFDI 27/1-460037581-BERD@NFDI. Esteban Garcés Arias sincerely thanks the Mentoring Program of the Faculty of Mathematics, Statistics and Informatics at LMU Munich, and the Munich Center for Machine Learning (MCML) for their support. Last but not least, we thank all the anonymous (meta) reviewers for their constructive feedback throughout the review process.

\bibliography{custom}

\appendix
\section*{Appendix}

\section{Discussion: Understanding the Interaction between VCM and Top-$p$}
\label{app:vcm and topp}

\subsection{The Truncation Paradox: Why does Top-$p$ lose its tail with VCM?}

As shown in our token migration analysis (Figure~\ref{fig:2}), integrating VCM generally shifts the probability mass from the absolute head (Rank 1--2) towards the middle tiers (Rank 3--12), mimicking human linguistic diversity. However, an apparent anomaly emerges when VCM is combined with Top-$p$ (nucleus sampling): the selection of tokens in the extreme long-tail (Rank $> 12$) drops from 8.9\% to 3.8\%.

One might superficially interpret this as VCM suppressing diversity when paired with Top-$p$. However, our subsequent analyses reveal that this phenomenon is not a limitation of VCM, but an inherent mathematical characteristic of the Top-$p$ cumulative probability mechanism when applied to a highly confident, sharpened distribution.

To precisely characterize this behavioral interaction, we designed two specific analyses focusing on the candidate pool size and threshold boundaries.

\subsection{Insight 1: VCM Induces Candidate Pool Collapse under Top-$p$}

\begin{figure*}[!ht]
    \centering
    \includegraphics[width=\linewidth]{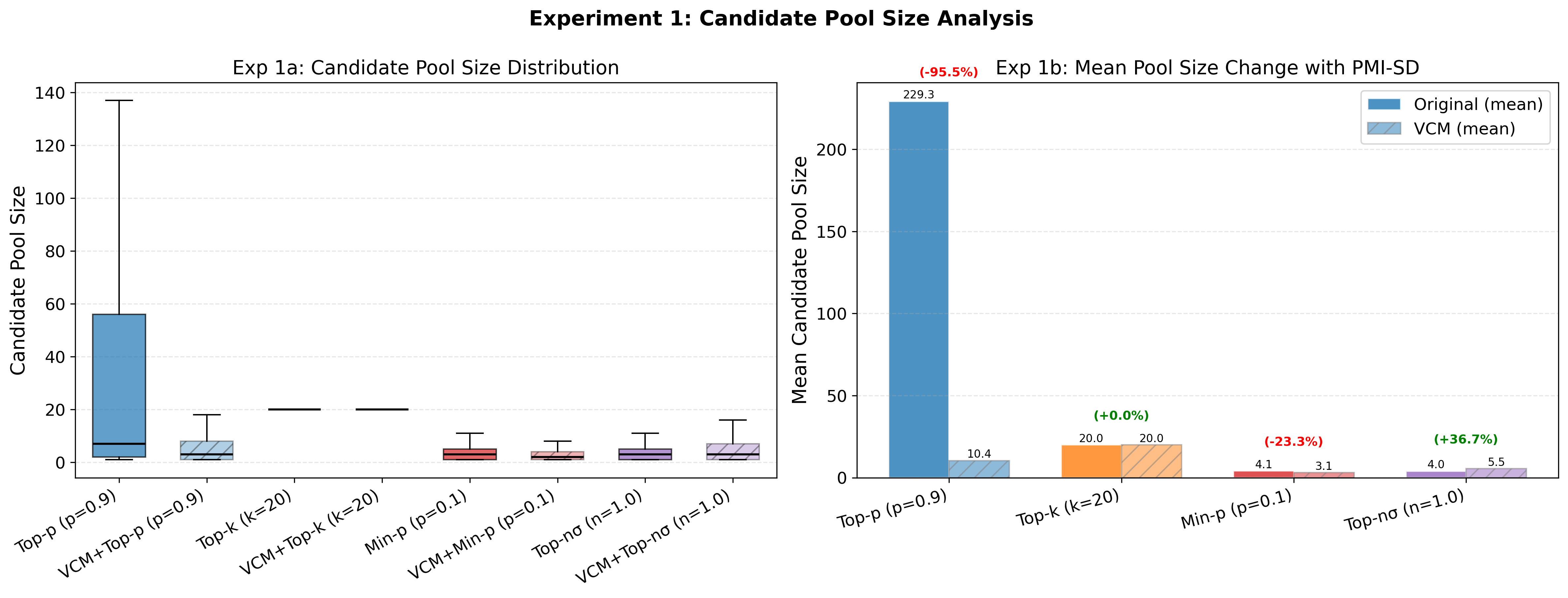}
    \caption{Candidate pool size comparison under different decoding strategies with and without VCM (Experiment~1).}
    \label{fig:topp-exp1}
\end{figure*}

In Experiment~1 (Figure~\ref{fig:topp-exp1}), we analyzed the candidate pool size, namely the number of tokens left after truncation. Strikingly, while other methods (Top-$k$, Min-$p$, Top-$n\sigma$) maintain or slightly increase their pool sizes when coupled with VCM, Top-$p$ experiences a substantial 95.5\% reduction in its mean candidate pool size, dropping from 229.3 to just 10.4 tokens.

\paragraph{Attribution.}
Top-$p$ operates by accumulating probability mass until it hits the threshold $p$. By amplifying context-aware tokens and strongly penalizing generic repetitions, VCM concentrates the probability mass into a smaller subset of highly relevant tokens, effectively making the distribution sharper. Consequently, the cumulative sum reaches the $p$ boundary much faster, forcing Top-$p$ to drastically shrink its candidate pool. When the pool size collapses to approximately $10$ tokens, it mathematically becomes impossible to sample tokens from Rank $> 12$, thus explaining the drop observed in Figure~\ref{fig:2}.

\subsection{Insight 2: Threshold Sweeping Confirms Mass Concentration}

\begin{figure*}[!ht]
    \centering
    \includegraphics[width=\linewidth]{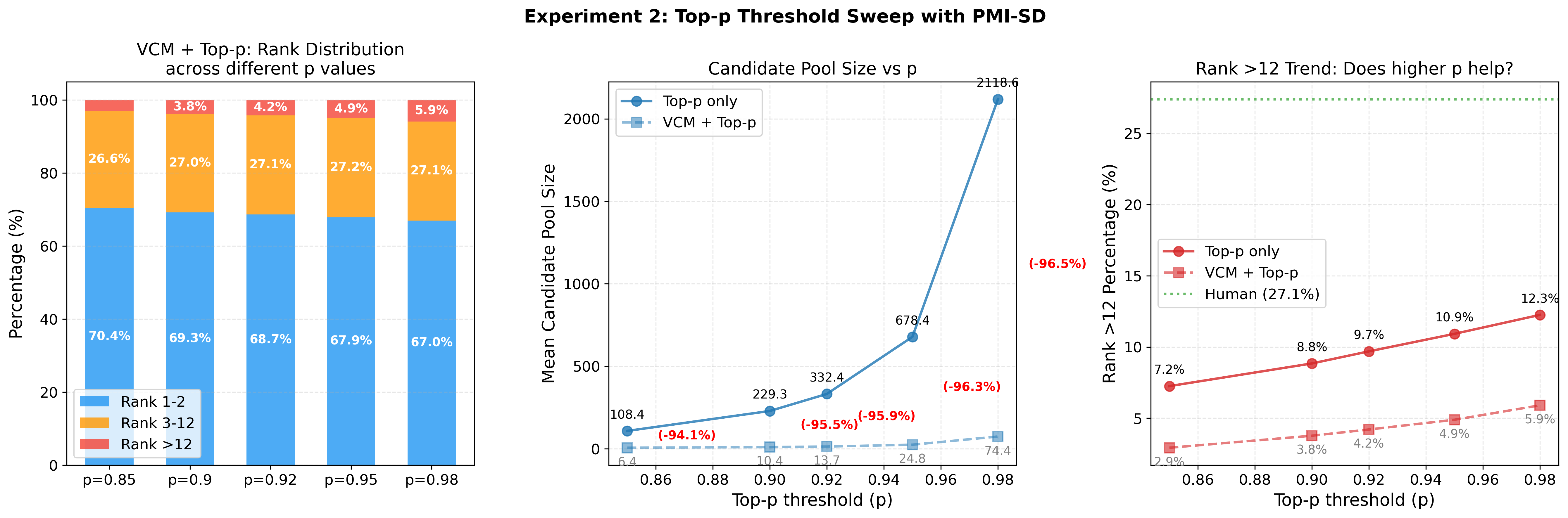}
    \caption{Threshold sweeping analysis of Top-$p$ with and without VCM (Experiment~2).}
    \label{fig:topp-exp2}
\end{figure*}

To verify whether this effect could be mitigated by relaxing the Top-$p$ threshold, we conducted Experiment~2 (Figure~\ref{fig:topp-exp2}). In standard Top-$p$ decoding, increasing $p$ from 0.85 to 0.98 causes an exponential explosion in the candidate pool size, from 104 to 2118.6 tokens, along with a steady increase in Rank $> 12$ tokens.

However, when VCM is applied, the candidate pool size remains stubbornly flat, barely reaching 74 tokens even at $p = 0.98$. Correspondingly, the percentage with Rank $> 12$ remains significantly compressed. This indicates that VCM reshapes the logits so that the large majority of the probability mass ($\geq 98\%$) is concentrated within roughly the top $70$ tokens.

\subsection{Overall Interpretation}

The combination of VCM and Top-$p$ is highly effective. It successfully pushes the generated tokens toward the human-like middle ranks (Rank 3--12). The suppression of Rank $> 12$ is merely an algorithmic artifact of Top-$p$ reacting to a higher-quality, more concentrated probability distribution. In essence, VCM resolves the model's uncertainty prior to sampling, rendering the wide, noisy safety net of traditional nucleus sampling largely redundant.


\section{Sensitivity Analysis of the Mixing Coefficient}
\label{app:alpha choice}

The mixing coefficient $\alpha$ serves as the primary hyperparameter in the VCM framework, governing the intervention strength. It dictates the extent to which the original logits are reshaped before downstream sampling algorithms are applied. A fundamental question arises regarding the robustness of VCM: \emph{Does the framework rely on a singularly tuned $\alpha$, or does it exhibit stable, generalizable operating regimes across different decoding backbones and diverse task paradigms?}

\begin{figure*}[!ht]
    \centering
    \includegraphics[width=\linewidth]{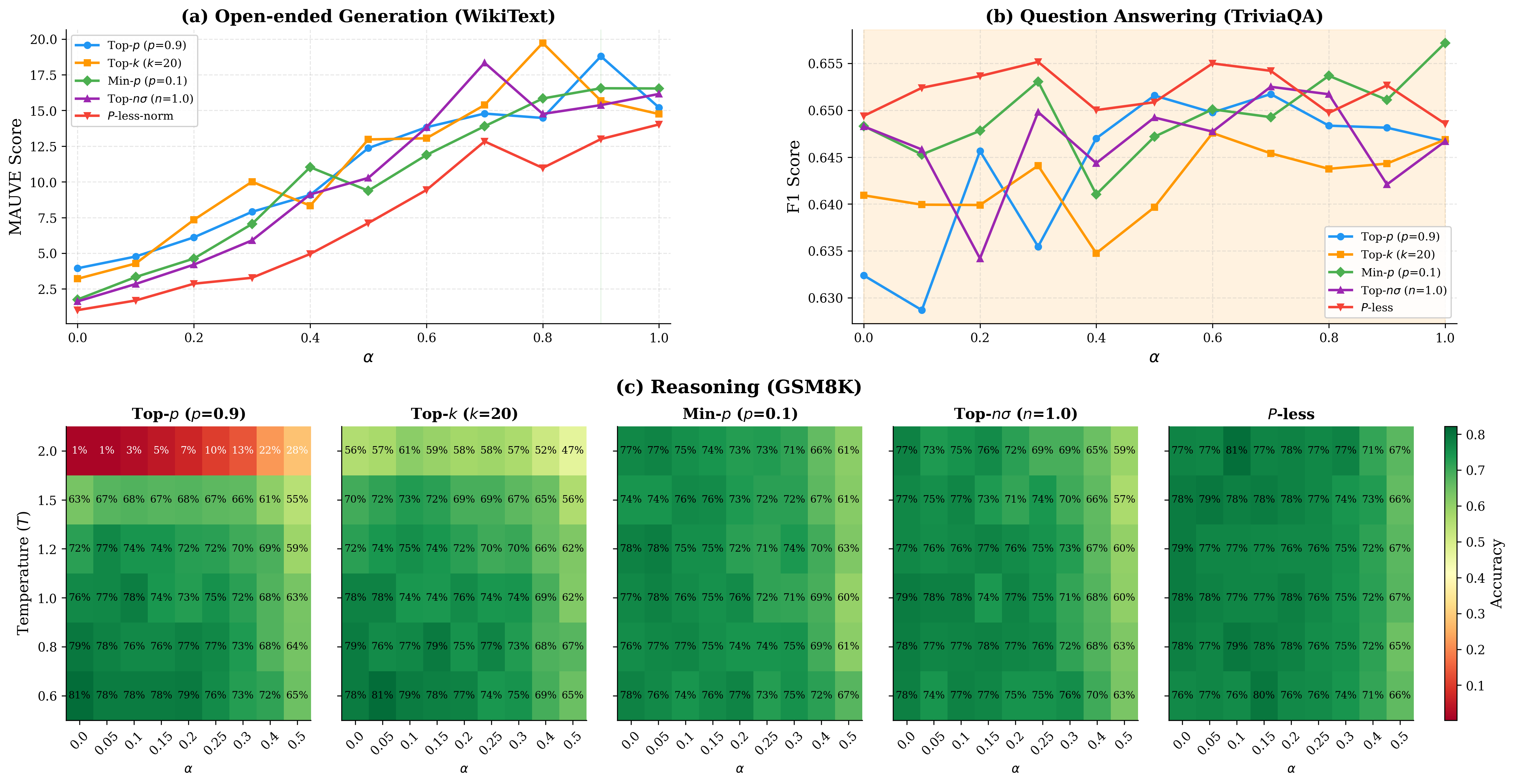}
    \caption{Sensitivity of VCM to the mixing coefficient $\alpha\in[0,1]$ across five samplers for (a) open-ended generation (WikiText, MAUVE), (b) factual QA (TriviaQA), and (c) mathematical reasoning (GSM8K). Each task shows a stable operating range rather than a single optimum.}
    \label{fig:alpha-analysis}
\end{figure*}

To systematically address this, we conduct comprehensive sensitivity analyses across three representative task families: (i) open-ended text generation, (ii) factual QA, and (iii) mathematical reasoning. For each task, we sweep $\alpha$ over the interval $[0.0, 1.0]$ ($[0.0, 0.5]$ for GSM8K) for all nine samplers; Figure~\ref{fig:alpha-analysis} shows five representative backbones (Top-$p$, Top-$k$, Min-$p$, Top-$n\sigma$, and $p$-less), and per-sampler recommendations are given in the accompanying tables. Overall, these findings indicate that VCM does not require brittle, dataset-specific tuning; instead, it exhibits stable, task-dependent operating ranges.

\begin{table*}[!ht]
\centering
\label{tab:alpha-wikitext}
\begin{tabular}{lccccc}
\hline
Method & Task Type & Dataset & \# Samples & Recommended &  $\alpha$ (Main Exp) \\
\hline
Top-$k$             & Open-ended Gen. & WikiText & $1{,}314$ & $[0.6, 0.8]$ & $0.6$ \\
Top-$p$             & Open-ended Gen. & WikiText & $1{,}314$ & $[0.6, 0.8]$ & $0.6$ \\
$\eta$-sampling     & Open-ended Gen. & WikiText & $1{,}314$ & $[0.6, 0.8]$ & $0.6$ \\
$\epsilon$-sampling & Open-ended Gen. & WikiText & $1{,}314$ & $[0.6, 0.8]$ & $0.6$ \\
Typical             & Open-ended Gen. & WikiText & $1{,}314$ & $[0.6, 0.8]$ & $0.6$ \\
Min-$p$             & Open-ended Gen. & WikiText & $1{,}314$ & $[0.6, 0.8]$ & $0.6$ \\
Top-$n\sigma$       & Open-ended Gen. & WikiText & $1{,}314$ & $[0.6, 0.8]$ & $0.6$ \\
$p$-less-norm       & Open-ended Gen. & WikiText & $1{,}314$ & $[0.6, 0.8]$ & $0.6$ \\
Min-$k$             & Open-ended Gen. & WikiText & $1{,}314$ & $[0.6, 0.8]$ & $0.6$ \\
\hline
\end{tabular}
\caption{Hyperparameter configurations and recommendations for open-ended generation.}
\end{table*}

\subsection{Open-ended Generation (WikiText)}

For open-ended generation, lexical diversity and the avoidance of repetitive loops are paramount. On the WikiText benchmark, the performance, measured via MAUVE score, which aligns closely with human judgments of text distribution, exhibits a distinct positive correlation with $\alpha$ once it departs from the near-zero regime. As observed in Figure~\ref{fig:alpha-analysis}(a), generation quality improves as $\alpha$ increases beyond the near-zero regime, with the strongest and most stable results in the medium-to-high range.

An elevated $\alpha$ aggressively penalizes globally frequent, generic tokens (via $z_{\text{prior}}$) while elevating context-specific vocabulary, successfully breaking the ``likelihood trap''. The interval $\alpha \in [0.6, 0.8]$ forms an empirically stable plateau across all evaluated samplers. Consequently, we adopt $\alpha = 0.6$ as a representative task-level configuration for our main experiments. When using models from the LLaMA family, a smaller value of $\alpha$ (e.g., $\alpha = 0.1$) is generally recommended, as it tends to yield better generation quality.

\begin{table*}[t]
\centering
\label{tab:alpha-triviaqa}
\begin{tabular}{lccccc}
\hline
Method & Task Type & Dataset & \# Samples & Recommended & $\alpha$ (Main Exp) \\
\hline
Top-$k$             & Truthfulness QA & TruthfulQA & $817$ & $[0.2, 0.5]$ & $0.4$ \\
Top-$p$             & Truthfulness QA & TruthfulQA & $817$ & $[0.2, 0.5]$ & $0.4$ \\
Typical             & Truthfulness QA & TruthfulQA & $817$ & $[0.2, 0.5]$ & $0.3$ \\
$\eta$-sampling     & Truthfulness QA & TruthfulQA & $817$ & $[0.2, 0.5]$ & $0.3$ \\
$\epsilon$-sampling & Truthfulness QA & TruthfulQA & $817$ & $[0.2, 0.5]$ & $0.4$ \\
Min-$p$             & Truthfulness QA & TruthfulQA & $817$ & $[0.2, 0.5]$ & $0.3$ \\
Top-$n\sigma$       & Truthfulness QA & TruthfulQA & $817$ & $[0.2, 0.5]$ & $0.3$ \\
$p$-less            & Truthfulness QA & TruthfulQA & $817$ & $[0.2, 0.5]$ & $0.4$ \\
Min-$k$             & Truthfulness QA & TruthfulQA & $817$ & $[0.1, 0.3]$ & $0.2$ \\
\hline
Top-$k$             & Factual QA & TriviaQA & $1{,}500$ & $[0.4, 0.6]$ & $0.5$ \\
Top-$p$             & Factual QA & TriviaQA & $1{,}500$ & $[0.4, 0.6]$ & $0.5$ \\
Typical             & Factual QA & TriviaQA & $1{,}500$ & $[0.4, 0.6]$ & $0.5$ \\
$\eta$-sampling     & Factual QA & TriviaQA & $1{,}500$ & $[0.4, 0.6]$ & $0.5$ \\
$\epsilon$-sampling & Factual QA & TriviaQA & $1{,}500$ & $[0.4, 0.6]$ & $0.5$ \\
Min-$p$             & Factual QA & TriviaQA & $1{,}500$ & $[0.4, 0.6]$ & $0.5$ \\
Top-$n\sigma$       & Factual QA & TriviaQA & $1{,}500$ & $[0.4, 0.6]$ & $0.5$ \\
$p$-less            & Factual QA & TriviaQA & $1{,}500$ & $[0.4, 0.6]$ & $0.5$ \\
Min-$k$             & Factual QA & TriviaQA & $1{,}500$ & $[0.4, 0.6]$ & $0.5$ \\
\hline
\end{tabular}
\caption{Hyperparameter configurations and recommendations for factual question answering.}
\end{table*}

\begin{table*}[!ht]
\centering
\label{tab:alpha-gsm8k}
\resizebox{\textwidth}{!}{
\begin{tabular}{lccccc}
\hline
Method & Task Type & Dataset & \# Samples & Recommended & $\alpha$ (Main Exp) \\
\hline
Top-$k$             & Reasoning & GSM8K & $1{,}319$ & $[0.01, 0.2]$ & $0.1$ ($T=1.0$), $0.2$ ($T=2.0$) \\
Top-$p$             & Reasoning & GSM8K & $1{,}319$ & $[0.01, 0.2]$ & $0.1$ ($T=1.0$), $0.2$ ($T=2.0$) \\
Typical             & Reasoning & GSM8K & $1{,}319$ & $[0.01, 0.2]$ & $0.1$ ($T=1.0$), $0.2$ ($T=2.0$) \\
$\eta$-sampling     & Reasoning & GSM8K & $1{,}319$ & $[0.01, 0.2]$ & $0.1$ ($T=1.0$), $0.2$ ($T=2.0$) \\
$\epsilon$-sampling & Reasoning & GSM8K & $1{,}319$ & $[0.01, 0.2]$ & $0.1$ ($T=1.0$), $0.2$ ($T=2.0$) \\
Min-$p$             & Reasoning & GSM8K & $1{,}319$ & $[0.01, 0.2]$ & $0.1$ ($T=1.0$), $0.1$ ($T=2.0$) \\
Top-$n\sigma$       & Reasoning & GSM8K & $1{,}319$ & $[0.01, 0.2]$ & $0.1$ ($T=1.0$), $0.01$ ($T=2.0$) \\
$p$-less            & Reasoning & GSM8K & $1{,}319$ & $[0.00, 0.01]$ & $0.005$ ($T=1.0$), $0.005$ ($T=2.0$) \\
Min-$k$             & Reasoning & GSM8K & $1{,}319$ & $[0.00, 0.1]$ & $0.1$ ($T=1.0$), $0.05$ ($T=2.0$) \\
\hline
Top-$k$             & Reasoning & MATH500 & $500$ & $[0.01, 0.6]$ & $0.1$ ($T=1.0$), $0.2$ ($T=2.0$) \\
Top-$p$             & Reasoning & MATH500 & $500$ & $[0.01, 0.6]$ & $0.1$ ($T=1.0$), $0.6$ ($T=2.0$) \\
Typical             & Reasoning & MATH500 & $500$ & $[0.01, 0.6]$ & $0.01$ ($T=1.0$), $0.5$ ($T=2.0$) \\
$\eta$-sampling     & Reasoning & MATH500 & $500$ & $[0.01, 0.6]$ & $0.1$ ($T=1.0$), $0.4$ ($T=2.0$) \\
$\epsilon$-sampling & Reasoning & MATH500 & $500$ & $[0.01, 0.2]$ & $0.1$ ($T=1.0$), $0.2$ ($T=2.0$) \\
Min-$p$             & Reasoning & MATH500 & $500$ & $[0.01, 0.2]$ & $0.1$ ($T=1.0$), $0.1$ ($T=2.0$) \\
Top-$n\sigma$       & Reasoning & MATH500 & $500$ & $[0.01, 0.1]$ & $0.1$ ($T=1.0$), $0.01$ ($T=2.0$) \\
$p$-less            & Reasoning & MATH500 & $500$ & $[0.01, 0.2]$ & $0.1$ ($T=1.0$), $0.2$ ($T=2.0$) \\
Min-$k$             & Reasoning & MATH500 & $500$ & $[0.00, 0.1]$ & $0.1$ ($T=1.0$), $0.001$ ($T=2.0$) \\

\hline
\end{tabular}
}
\caption{Hyperparameter configurations and recommendations for mathematical reasoning.}
\end{table*}

\subsection{Factual Question Answering (TriviaQA)}

Factual QA inherently diverges from open-ended generation; it heavily penalizes unconstrained lexical exploration and hallucinatory behaviors. Evaluating on the TriviaQA benchmark, we observe a more moderated sensitivity to $\alpha$ (Figure~\ref{fig:alpha-analysis}(b)).

While the performance curves are relatively flatter, a discernible preference for intermediate values emerges. In this setting, an intermediate $\alpha$ provides sufficient contextual sharpening, suppressing vague, generic fallback answers, without introducing excessive variance that might distort factual entities. Across samplers, the strongest performance generally falls within $\alpha\in[0.4,0.6]$. To balance factual precision and contextual focus, we select $\alpha=0.5$ for the main evaluations.

\subsection{Mathematical Reasoning (GSM8K)}

Mathematical reasoning benchmarks, such as GSM8K, demand exact step-by-step logic and highly sensitive token-level dependencies. As depicted in the heatmaps of Figure~\ref{fig:alpha-analysis}(c), GSM8K reveals a highly structured, conservative preference regime.

Across all sampling algorithms and temperature settings, accuracy peaks exclusively in the low-$\alpha$ domain, specifically clustering around $\alpha \in [0.01, 0.2]$. This phenomenon implies that rigid logical chains are susceptible to disruption by aggressive logits modulation. A very large $\alpha$ risks forcefully re-routing an otherwise coherent mathematical thought process. Therefore, VCM functions best as a subtle ``nudge'' in reasoning tasks. Aligning with these observations, we deploy highly conservative values in our main experiments, adjusting slightly based on the base temperature.

\subsection{Overall Discussion}

Synthesizing the empirical evidence across these three distinct paradigms, we conclude that there is no singular ``silver bullet'' value for $\alpha$, nor should there be. The optimal strength of intervention is fundamentally correlated with the cognitive requirements of the task:

     \textbf{High Creativity \& Diversity (WikiText):} benefits from aggressive modulation ($\alpha \approx 0.6$).
     
     \textbf{Factual Recall (TriviaQA):} requires balanced, moderate intervention ($\alpha \approx 0.5$).
     
     \textbf{Strict Logic \& Reasoning (GSM8K):} demands highly conservative, subtle modulation ($\alpha \leq 0.2$).

Crucially, within each task category, VCM exhibits stable performance across a broad interval of $\alpha$ and across all evaluated samplers, rather than depending on a single finely tuned value.


\section{Ablation Study}
\label{sec:ablation}

To comprehensively validate the design choices and the robustness of the VCM framework, we conduct detailed ablation studies focusing on three critical aspects: the synergy of its orthogonal components, the choice of the unconditional prior, and the sensitivity of the context window size.

\subsection{Analyzing the Synergy of Contextual Searchlight and Adaptive Self-Debiasing}
\label{sec:ablation_components}

VCM is constructed upon two core mechanisms: Contextual Searchlight ($\mathcal{S}_{\text{prior}}$) and Adaptive Self-Debiasing ($\mathcal{C}_{\text{rep}}$). To understand their individual contributions, we decouple the VCM signal $\mathcal{I}_t$ and evaluate each component independently against the Top-$p$ baseline. Results across WikiText, TriviaQA, and GSM8K are reported in Table \ref{tab:ablation_components}.

\begin{table*}[htbp]
\centering
\resizebox{0.9\textwidth}{!}{
\begin{tabular}{lccccc}
\toprule
\multirow{2}{*}{\textbf{Condition}} & \multicolumn{2}{c}{\textbf{WikiText}} & \multicolumn{2}{c}{\textbf{TriviaQA}} & \textbf{GSM8K} \\
\cmidrule(lr){2-3} \cmidrule(lr){4-5} \cmidrule(lr){6-6}
 & MAUVE $\uparrow$ & BERTScore $\uparrow$ & EM $\uparrow$ & F1 $\uparrow$ & EM ($T=1.0$) $\uparrow$ \\
\midrule
Baseline (Top-$p$ only) & 2.81 & 42.80 & 49.80 & 62.96 & 73.92 \\
+ $\mathcal{S}_{\text{prior}}$ only (PMI) & 1.75 & 43.74 & 44.93 & 61.42 & 74.68 \\
+ $\mathcal{C}_{\text{rep}}$ only (Adaptive Rep) & \textbf{24.11} & \textbf{50.04} & 50.13 & 61.29 & 73.77 \\
\midrule
\textbf{+ VCM ($\mathcal{S}_{\text{prior}} + \mathcal{C}_{\text{rep}}$)} & 14.71 & 46.97 & \textbf{52.93} & \textbf{64.63} & \textbf{76.65} \\
\bottomrule
\end{tabular}
}
\caption{Ablation of VCM components. The full VCM achieves the optimal balance between generation diversity (MAUVE), factual recall (TriviaQA), and reasoning (GSM8K), avoiding the severe performance drops observed when using either component in isolation. \textbf{Bold} marks the best value per column.}
\label{tab:ablation_components}
\end{table*}

\paragraph{Analysis.} The empirical results strongly validate that $\mathcal{S}_{\text{prior}}$ and $\mathcal{C}_{\text{rep}}$ act as a highly effective, orthogonal ``push-pull'' mechanism:
\begin{itemize}
    \item \textbf{$\mathcal{S}_{\text{prior}}$ Only:} Relying solely on the contextual prior improves mathematical reasoning (GSM8K EM increases to $74.68$) by elevating context-specific tokens. However, without adaptive debiasing, it severely collapses generative diversity (WikiText MAUVE drops from $2.81$ to $1.75$) and hurts factual recall (TriviaQA EM drops from $49.80$ to $44.93$).
    \item \textbf{$\mathcal{C}_{\text{rep}}$ Only:} Conversely, deploying only the adaptive repetition penalty yields the highest diversity (WikiText MAUVE rises to $24.11$). Yet it slightly reduces reasoning accuracy (GSM8K EM $73.92\rightarrow73.77$), consistent with the fact that mathematical reasoning requires correctly regenerating repeated symbols (e.g., ``='', digits, operators) that an unconstrained repetition penalty may suppress.
    \item \textbf{Full VCM:} When integrated, the complete VCM mathematically balances these forces. $\mathcal{S}_{\text{prior}}$ safeguards the generation of critical, context-relevant tokens (preventing $\mathcal{C}_{\text{rep}}$ from erroneously penalizing necessary repetitions like math symbols), while $\mathcal{C}_{\text{rep}}$ mitigates the likelihood trap and repetitive loops. This synergy yields the most balanced profile across the three axes: it improves factual recall (TriviaQA EM $52.93$) and reasoning accuracy (GSM8K $76.65$) over the baseline while retaining strong diversity (MAUVE $14.71$).
\end{itemize}

\subsection{Sensitivity to Unconditional Prior}
\label{sec:ablation_prior}

The contextual component of VCM relies on a pre-computed reference logit vector $\mathbf{z}_{\mathrm{prior}}$. To examine how strongly the method depends on this choice, we evaluate six prior constructions on 500-sample subsets of WikiText and GSM8K. In our implementation, an empty string produces the same prior as the BOS token and is therefore not listed separately.

Table~\ref{tab:prior_ablation} shows that no single prior dominates across tasks and metrics. For example, the neutral prompt ``Answer:'' improves GSM8K exact match relative to the BOS prior, but substantially reduces WikiText MAUVE and BERTScore. Similarly, other alternatives favor different aspects of generation quality or reasoning accuracy. These results suggest that task-format-specific priors can introduce benchmark-specific preferences rather than providing a uniformly better reference distribution.

\begin{table*}[!ht]
\centering
\begin{tabular}{lcccc c}
\toprule
& \multicolumn{4}{c}{\textbf{WikiText}}
& \textbf{GSM8K} \\
\cmidrule(lr){2-5}
\cmidrule(lr){6-6}
\textbf{Prior}
& \textbf{MAUVE} $\uparrow$
& \textbf{BERTScore} $\uparrow$
& \textbf{Rep-2} $\downarrow$
& \textbf{Distinct-2} $\uparrow$
& \textbf{EM} $\uparrow$ \\
\midrule
BOS (default)
& 17.30
& 46.63
& 61.35
& 25.87
& 74.2 \\

PAD
& 18.90
& \textbf{47.07}
& 59.93
& 24.39
& 74.6 \\

Common word (``the'')
& \textbf{19.77}
& 46.80
& 60.28
& 25.83
& \textbf{77.4} \\

Newline
& 17.80
& 46.41
& 62.00
& 25.15
& 75.2 \\

Neutral prompt (``Answer:'')
& 14.11
& 45.39
& 59.84
& 27.63
& 77.2 \\

Averaged neutral prior
& 19.73
& 46.23
& \textbf{57.92}
& \textbf{27.94}
& 76.0 \\
\bottomrule
\end{tabular}
\caption{Sensitivity to the choice of $\mathbf{z}_{\mathrm{prior}}$ on 500-sample subsets of WikiText and GSM8K. All variants are evaluated with a Top-$p$ sampling backbone ($p=0.9$, $T=1.0$), using $\alpha=0.6$ for WikiText and $\alpha=0.1$ for GSM8K. No single prior dominates across tasks and metrics. We retain BOS as the default because it is semantically minimal and task-agnostic, and requires neither prompt engineering nor corpus-level estimation.}
\label{tab:prior_ablation}
\end{table*}

\begin{table*}[!ht]
\centering
\resizebox{0.9\textwidth}{!}{
\begin{tabular}{lccccc}
\toprule
\multirow{2}{*}{\textbf{Window Size ($W$)}} & \multicolumn{2}{c}{\textbf{WikiText}} & \multicolumn{2}{c}{\textbf{TriviaQA}} & \textbf{GSM8K} \\
\cmidrule(lr){2-3} \cmidrule(lr){4-5} \cmidrule(lr){6-6}
 & MAUVE $\uparrow$ & BERTScore $\uparrow$ & EM $\uparrow$ & F1 $\uparrow$ & EM ($T=1.0$) $\uparrow$ \\
\midrule
16 & 2.28 & 44.57 & 46.93 & 61.75 & 74.07 \\
32 & 3.34 & 45.18 & 47.40 & 62.34 & 75.97 \\
64 & 7.46 & 46.05 & 48.67 & 62.83 & 76.19 \\
\midrule
\textbf{128} (Default) & 14.71 & 46.97 & 52.93 & \textbf{64.63} & \textbf{76.65} \\
\midrule
256 & \textbf{16.47} & 48.03 & \textbf{53.87} & 64.50 & 76.04 \\
512 & 16.31 & \textbf{48.07} & 53.07 & 63.55 & 75.21 \\
\bottomrule
\end{tabular}
}
\caption{Impact of the context window size ($W$) on the adaptive repetition penalty. $W=128$ offers the best overall balance, mitigating short-term loops while preserving long-term semantic coherence across diverse tasks.}
\label{tab:ablation_window}
\end{table*}

We therefore retain BOS as the default prior, not because it is universally optimal, but because it provides a semantically minimal and task-agnostic reference without requiring prompt engineering or corpus-level estimation. Learned or task-adaptive prior construction remains an interesting direction for future work.

\subsection{Influence of Context Window Size}
\label{sec:ablation_window}

The adaptive repetition component $\mathcal{C}_{\text{rep}}$ relies on a local frequency count within a moving window of size $W$. We sweep $W \in \{16, 32, 64, 128, 256, 512\}$ to locate the optimal historical scope for penalization. 

\paragraph{Analysis.} Table \ref{tab:ablation_window} reveals a distinct inverted-U performance curve, highlighting a critical trade-off between recency bias and long-term semantic coherence:

\begin{table*}[!ht]
\centering
\begin{tabular}{llccc}
\toprule
\textbf{Benchmark}
& \textbf{Sampler}
& \textbf{Baseline length}
& \textbf{$+$VCM length}
& \textbf{Relative change} \\
& & \textbf{(mean $\pm$ std)}
& \textbf{(mean $\pm$ std)}
& \\
\midrule
WikiText
& Top-$p$ / Top-$k$ / Min-$p$
& $256.00 \pm 0.00$
& $256.00 \pm 0.00$
& $0.0\%$ \\
\midrule
TriviaQA
& Top-$p$
& $143.98 \pm 72.73$
& $122.01 \pm 56.34$
& $-15.3\%$ \\

& Top-$k$
& $142.50 \pm 65.83$
& $114.18 \pm 51.24$
& $-19.9\%$ \\

& Min-$p$
& $138.50 \pm 58.44$
& $122.28 \pm 56.95$
& $-11.7\%$ \\
\midrule
GSM8K
& Top-$p$
& $115.13 \pm 44.64$
& $118.14 \pm 44.37$
& $+2.6\%$ \\

& Top-$k$
& $110.49 \pm 38.75$
& $117.72 \pm 47.05$
& $+6.5\%$ \\

& Min-$p$
& $113.93 \pm 47.09$
& $120.18 \pm 43.39$
& $+5.5\%$ \\
\bottomrule
\end{tabular}
\caption{Generation lengths under Top-$p$, Top-$k$, and Min-$p$. Lengths are reported as mean $\pm$ standard deviation. WikiText uses a fixed 256-token generation budget; VCM shortens generations on TriviaQA and causes only moderate length changes on GSM8K.}
\label{tab:generation_length}
\end{table*}

\begin{itemize}
    \item \textbf{Too small ($W \leq 64$):} The model suffers from ``short-sightedness.'' It fails to penalize macro-level repetitive loops, resulting in severely degraded open-ended text quality (WikiText MAUVE $\leq 7.46$) and sub-optimal reasoning.
    \item \textbf{Too large ($W \geq 256$):} While an expansive window marginally improves WikiText diversity, it yields no further gains on TriviaQA F1 and reduces GSM8K accuracy. A very large window indiscriminately penalizes necessary topic-specific entities or mathematical operands that naturally re-appear across a long reasoning chain, causing performance drops on GSM8K and, at $W=512$, also on TriviaQA.
    \item \textbf{The Sweet Spot ($W=128$):} This intermediate window size optimally calibrates the memory span. It effectively disrupts local degenerative loops while safely ``forgetting'' distant tokens, allowing context-critical entities to be regenerated when logically necessary. 
    Consequently, we universally adopt $W=128$ as the robust default across all task families in our main experiments.
\end{itemize}

\section{Generation Length Analysis}
\label{app:generation_length}

To examine whether the performance gains of VCM are associated with systematically longer generations, we report the mean generation length under Top-$p$, Top-$k$, and Min-$p$ on WikiText, TriviaQA, and GSM8K. The corresponding maximum generation budgets are 256, 512, and 1024 tokens, respectively.

As shown in Table~\ref{tab:generation_length}, all WikiText generations reach the fixed 256-token budget by design. On TriviaQA, VCM produces shorter outputs under all three samplers, with mean length reductions between 11.7\% and 19.9\%. On GSM8K, the mean length increases only moderately, ranging from 2.6\% to 6.5\%. All outputs terminate according to the configured stopping or EOS conditions. Overall, the improvements obtained with VCM are not attributable to systematically longer generations.

\onecolumn

\section{LLM-as-a-Judge Results}
\label{app:llmjudge}
\subsection{Prompt}
We use the following prompt template for judge-based evaluation:

\begin{quote}
\small
\textbf{System instruction:} \\
\texttt{You are a precise text quality evaluator. Always respond with valid JSON only.}

\vspace{0.5em}
\textbf{User prompt:} \\
\texttt{You are an expert text quality evaluator. You will be given a prefix text (context), a reference text (ideal continuation), and a candidate text to evaluate.}

\vspace{0.5em}
\texttt{Prefix (Context): \{prefix\_text\}}

\texttt{Reference (Ideal Continuation): \{reference\_text\}}

\texttt{Candidate Text to Evaluate: \{candidate\_text\}}

\vspace{0.5em}
\texttt{Score the candidate text on each dimension from 1 to 5, where 1 = Very Poor and 5 = Excellent.}

\texttt{Dimensions:}
\begin{itemize}
    \item \texttt{Coherence: Does the text logically follow the prefix? Is it internally consistent?}
    \item \texttt{Fluency: Is the text grammatically correct and natural?}
    \item \texttt{Relevance: Is the content topically aligned with the prefix?}
    \item \texttt{Faithfulness: How well does it match the reference in meaning?}
    \item \texttt{Informativeness: Does it convey rich, meaningful content?}
\end{itemize}

\texttt{Required Output Format (strict JSON):}
\begin{verbatim}
{
  "scores": {
    "coherence": <int 1-5>,
    "fluency": <int 1-5>,
    "relevance": <int 1-5>,
    "faithfulness": <int 1-5>,
    "informativeness": <int 1-5>
  },
  "overall": <float, weighted average>,
  "justification": "Brief explanation for the scores"
}
\end{verbatim}

\texttt{Return ONLY the JSON object, no additional text.}
\end{quote}
\clearpage

\subsection{Results}

To ensure the reliability and statistical significance of our LLM-as-a-Judge evaluations, this section provides comprehensive comparative results across all decoding baselines. Tables \ref{tab:topkllmjudge} through \ref{tab:minkllmjudge} detail the win rates, tie rates, and fine-grained dimension scores for both DeepSeek V3.2 and Qwen3.6-Plus. Furthermore, we report Cohen’s $\kappa$ and overall agreement percentages to demonstrate the consistency of the evaluators. Inter-judge agreement is fair to moderate (Cohen's $\kappa$ between $0.15$ and $0.40$ across settings), and both judges prefer VCM outputs over their baselines, indicating that the observed gains are consistent across two independent evaluators rather than specific to a single judge.
\begin{table*}[htbp]
\centering
\resizebox{\textwidth}{!}{
\begin{tabular}{lcccccc|cccccc|cc}
\hline
\multirow{2}{*}{Dimension} 
& \multicolumn{6}{c|}{DeepSeek V3.2} 
& \multicolumn{6}{c|}{Qwen3.6-Plus} 
& \multicolumn{2}{c}{Cohen's $\kappa$ \& Agree.} \\
\cline{2-15}
& Avg(0) & Avg(V) & $\Delta$ & 0 Wins & V Wins & Tie
& Avg(0) & Avg(V) & $\Delta$ & 0 Wins & V Wins & Tie
& $\kappa$ & Agree\%  \\
\hline
coherence         & 1.060 & 1.330 & +0.270 & 5  & 32 & 63  & 1.030 & 1.071 & +0.041 & 3  & 6  & 91  & 0.261 & 70.0\%  \\
fluency           & 1.570 & 2.170 & +0.600 & 15 & 51 & 34  & 1.300 & 1.758 & +0.458 & 14 & 47 & 39  & 0.374 & 62.0\%  \\
relevance         & 2.150 & 2.150 & +0.000 & 24 & 26 & 50  & 1.540 & 1.657 & +0.117 & 18 & 27 & 55  & 0.117 & 46.0\%  \\
faithfulness      & 1.000 & 1.000 & +0.000 & 0  & 0  & 100 & 1.010 & 1.000 & -0.010 & 2  & 0  & 98  & 0.000 & 98.0\%  \\
informativeness   & 1.020 & 1.260 & +0.240 & 2  & 24 & 74  & 1.000 & 1.020 & +0.020 & 1  & 2  & 97  & 0.098 & 75.0\%  \\
\hline
OVERALL & 1.333 & 1.527 & +0.194 & 24 & 59 & 17  & 1.163 & 1.266 & +0.103 & 26 & 53 & 21  & 0.236 & 55.0\%  \\
\hline
\end{tabular}
}
\caption{Top-$k$ and + VCM results. Comparison of DeepSeek V3.2 and Qwen3.6-Plus with Cohen's $\kappa$ and agreement statistics. Here, $0$ denotes the baseline sampler and $V$ denotes +VCM.}
\label{tab:topkllmjudge}
\end{table*}

\begin{table*}[!htbp]
\centering
\resizebox{\textwidth}{!}{%
\begin{tabular}{lcccccc|cccccc|cc}
\hline
\multirow{2}{*}{Dimension}
& \multicolumn{6}{c|}{DeepSeek V3.2}
& \multicolumn{6}{c|}{Qwen3.6-Plus}
& \multicolumn{2}{c}{Cohen's $\kappa$ \& Agree.} \\
\cline{2-15}
& Avg(0) & Avg(V) & $\Delta$ & 0 Wins & V Wins & Tie
& Avg(0) & Avg(V) & $\Delta$ & 0 Wins & V Wins & Tie
& $\kappa$ & Agree\%  \\
\hline
coherence          & 1.080 & 1.240 & +0.160 & 6  & 18 & 76  & 1.000 & 1.120 & +0.120 & 0  & 12 & 88  & 0.289 & 78.0\%  \\
fluency            & 1.530 & 2.080 & +0.550 & 15 & 45 & 40  & 1.242 & 1.780 & +0.538 & 9  & 43 & 48  & 0.368 & 62.0\%  \\
relevance          & 2.130 & 2.180 & +0.050 & 23 & 26 & 51  & 1.556 & 1.580 & +0.024 & 24 & 26 & 50  & 0.229 & 52.0\% \\
faithfulness       & 1.000 & 1.000 & +0.000 & 0  & 0  & 100 & 1.000 & 1.010 & +0.010 & 0  & 2  & 98  & 0.000 & 98.0\%  \\
informativeness    & 1.020 & 1.240 & +0.220 & 2  & 21 & 77  & 1.000 & 1.050 & +0.050 & 0  & 6  & 94  & 0.279 & 81.0\%  \\
\hline
OVERALL & 1.329 & 1.494 & +0.165 & 26 & 56 & 18  & 1.147 & 1.273 & +0.126 & 22 & 51 & 27  & 0.326 & 59.0\%  \\
\hline
\end{tabular}%
}
\caption{Top-$p$ and + VCM results. Comparison of DeepSeek V3.2 and Qwen3.6-Plus with Cohen's $\kappa$ and agreement statistics. Here, $0$ denotes the baseline sampler and $V$ denotes +VCM.}
\label{tab:toppllmjudge}
\end{table*}

\begin{table*}[!htbp]
\centering
\resizebox{\textwidth}{!}{%
\begin{tabular}{lcccccc|cccccc|cc}
\hline
\multirow{2}{*}{Dimension}
& \multicolumn{6}{c|}{DeepSeek V3.2}
& \multicolumn{6}{c|}{Qwen3.6-Plus}
& \multicolumn{2}{c}{Cohen's $\kappa$ \& Agree.} \\
\cline{2-15}
& Avg(0) & Avg(V) & $\Delta$ & 0 Wins & V Wins & Tie
& Avg(0) & Avg(V) & $\Delta$ & 0 Wins & V Wins & Tie
& $\kappa$ & Agree\%  \\
\hline
coherence          & 1.110 & 1.360 & +0.250 & 4  & 29 & 67  & 1.020 & 1.100 & +0.080 & 0  & 8  & 92  & 0.140 & 69.0\%  \\
fluency            & 1.530 & 2.040 & +0.510 & 11 & 46 & 43  & 1.340 & 1.800 & +0.460 & 7  & 37 & 56  & 0.243 & 56.0\%  \\
relevance          & 2.110 & 2.180 & +0.070 & 24 & 28 & 48  & 1.570 & 1.650 & +0.080 & 25 & 29 & 46  & 0.279 & 54.0\%  \\
faithfulness       & 1.000 & 1.000 & +0.000 & 0  & 0  & 100 & 1.000 & 1.010 & +0.010 & 0  & 1  & 99  & 0.000 & 99.0\%  \\
informativeness    & 1.040 & 1.290 & +0.250 & 2  & 25 & 73  & 1.000 & 1.010 & +0.010 & 0  & 1  & 99  & -0.019 & 72.0\%  \\
\hline
OVERALL  & 1.335 & 1.526 & +0.191 & 21 & 59 & 20  & 1.170 & 1.279 & +0.109 & 23 & 46 & 31  & 0.240 & 53.0\%  \\
\hline
\end{tabular}%
}
\caption{$\eta$-sampling and + VCM results. Comparison of DeepSeek V3.2 and Qwen3.6-Plus with Cohen's $\kappa$ and agreement statistics. Here, $0$ denotes the baseline sampler and $V$ denotes +VCM.}
\label{tab:etallmjudge}
\end{table*}

\begin{table*}[!htbp]
\centering
\resizebox{\textwidth}{!}{%
\begin{tabular}{lcccccc|cccccc|cc}
\hline
\multirow{2}{*}{Dimension}
& \multicolumn{6}{c|}{DeepSeek V3.2}
& \multicolumn{6}{c|}{Qwen3.6-Plus}
& \multicolumn{2}{c}{Cohen's $\kappa$ \& Agree.} \\
\cline{2-15}
& Avg(0) & Avg(V) & $\Delta$ & 0 Wins & V Wins & Tie
& Avg(0) & Avg(V) & $\Delta$ & 0 Wins & V Wins & Tie
& $\kappa$ & Agree\%  \\
\hline
coherence          & 1.070 & 1.320 & +0.250 & 3  & 28 & 69  & 1.020 & 1.101 & +0.081 & 1  & 8  & 91  & 0.313 & 76.0\%  \\
fluency            & 1.500 & 2.010 & +0.510 & 13 & 47 & 40  & 1.340 & 1.737 & +0.397 & 15 & 43 & 42  & 0.427 & 65.0\%  \\
relevance          & 2.200 & 2.210 & +0.010 & 26 & 26 & 48  & 1.560 & 1.697 & +0.137 & 18 & 29 & 53  & 0.310 & 57.0\%  \\
faithfulness       & 1.000 & 1.000 & +0.000 & 0  & 0  & 100 & 1.000 & 1.000 & +0.000 & 1  & 0  & 99  & 0.000 & 99.0\%  \\
informativeness    & 1.030 & 1.250 & +0.220 & 2  & 23 & 75  & 1.000 & 1.030 & +0.030 & 1  & 3  & 96  & 0.084 & 75.0\%  \\
\hline
OVERALL  & 1.337 & 1.511 & +0.174 & 26 & 59 & 15  & 1.168 & 1.280 & +0.112 & 24 & 52 & 24  & 0.361 & 62.0\%  \\
\hline
\end{tabular}%
}
\caption{$\epsilon$-sampling and + VCM results. Comparison of DeepSeek V3.2 and Qwen3.6-Plus with Cohen's $\kappa$ and agreement statistics. Here, $0$ denotes the baseline sampler and $V$ denotes +VCM.}
\label{tab:epsilonllmjudge}
\end{table*}

\begin{table*}[!htbp]
\centering
\resizebox{\textwidth}{!}{%
\begin{tabular}{lcccccc|cccccc|cc}
\hline
\multirow{2}{*}{Dimension}
& \multicolumn{6}{c|}{DeepSeek V3.2}
& \multicolumn{6}{c|}{Qwen3.6-Plus}
& \multicolumn{2}{c}{Cohen's $\kappa$ \& Agree.} \\
\cline{2-15}
& Avg(0) & Avg(V) & $\Delta$ & 0 Wins & V Wins & Tie
& Avg(0) & Avg(V) & $\Delta$ & 0 Wins & V Wins & Tie
& $\kappa$ & Agree\%  \\
\hline
coherence          & 1.070 & 1.330 & +0.260 & 2  & 28 & 70  & 1.010 & 1.110 & +0.100 & 0  & 11 & 89  & 0.307 & 76.0\%  \\
fluency            & 1.420 & 2.010 & +0.590 & 11 & 51 & 38  & 1.242 & 1.650 & +0.408 & 6  & 39 & 55  & 0.249 & 56.0\%  \\
relevance          & 2.160 & 2.170 & +0.010 & 25 & 25 & 50  & 1.475 & 1.720 & +0.245 & 12 & 35 & 53  & 0.158 & 48.0\%  \\
faithfulness       & 1.000 & 1.000 & +0.000 & 0  & 0  & 100 & 1.010 & 1.000 & -0.010 & 1  & 1  & 98  & 0.000 & 98.0\%  \\
informativeness    & 1.010 & 1.200 & +0.190 & 0  & 18 & 82  & 1.000 & 1.030 & +0.030 & 0  & 4  & 96  & 0.222 & 84.0\%  \\
\hline
OVERALL  & 1.314 & 1.498 & +0.184 & 19 & 62 & 19  & 1.136 & 1.274 & +0.138 & 16 & 55 & 29  & 0.146 & 51.0\%  \\
\hline
\end{tabular}%
}
\caption{Typical sampling and + VCM results. Comparison of DeepSeek V3.2 and Qwen3.6-Plus with Cohen's $\kappa$ and agreement statistics. Here, $0$ denotes the baseline sampler and $V$ denotes +VCM.}
\label{tab:typicalllmjudge}
\end{table*}

\begin{table*}[!htbp]
\centering
\resizebox{\textwidth}{!}{%
\begin{tabular}{lcccccc|cccccc|cc}
\hline
\multirow{2}{*}{Dimension}
& \multicolumn{6}{c|}{DeepSeek V3.2}
& \multicolumn{6}{c|}{Qwen3.6-Plus}
& \multicolumn{2}{c}{Cohen's $\kappa$ \& Agree.} \\
\cline{2-15}
& Avg(0) & Avg(V) & $\Delta$ & 0 Wins & V Wins & Tie
& Avg(0) & Avg(V) & $\Delta$ & 0 Wins & V Wins & Tie
& $\kappa$ & Agree\%  \\
\hline
coherence          & 1.050 & 1.350 & +0.300 & 1  & 31 & 68  & 1.030 & 1.070 & +0.040 & 2  & 6  & 92  & 0.297 & 75.0\%  \\
fluency            & 1.420 & 1.870 & +0.450 & 15 & 49 & 36  & 1.290 & 1.540 & +0.250 & 10 & 34 & 56  & 0.319 & 58.0\%  \\
relevance          & 2.220 & 2.360 & +0.140 & 17 & 28 & 55  & 1.590 & 1.740 & +0.150 & 23 & 32 & 45  & 0.231 & 52.0\%  \\
faithfulness       & 1.000 & 1.000 & +0.000 & 0  & 0  & 100 & 1.000 & 1.000 & +0.000 & 0  & 0  & 100 & 1.000 & 100.0\%  \\
informativeness    & 1.020 & 1.150 & +0.130 & 1  & 13 & 86  & 1.020 & 1.010 & -0.010 & 2  & 1  & 97  & 0.148 & 86.0\%  \\
\hline
OVERALL  & 1.323 & 1.513 & +0.190 & 23 & 59 & 18  & 1.172 & 1.248 & +0.076 & 28 & 46 & 26  & 0.287 & 56.0\%  \\
\hline
\end{tabular}%
}
\caption{Min-$p$ sampling and + VCM results. Comparison of DeepSeek V3.2 and Qwen3.6-Plus with Cohen's $\kappa$ and agreement statistics. Here, $0$ denotes the baseline sampler and $V$ denotes +VCM.}
\label{tab:minpllmjudge}
\end{table*}

\begin{table*}[!htbp]
\centering
\resizebox{\textwidth}{!}{%
\begin{tabular}{lcccccc|cccccc|cc}
\hline
\multirow{2}{*}{Dimension}
& \multicolumn{6}{c|}{DeepSeek V3.2}
& \multicolumn{6}{c|}{Qwen3.6-Plus}
& \multicolumn{2}{c}{Cohen's $\kappa$ \& Agree.} \\
\cline{2-15}
& Avg(0) & Avg(V) & $\Delta$ & 0 Wins & V Wins & Tie
& Avg(0) & Avg(V) & $\Delta$ & 0 Wins & V Wins & Tie
& $\kappa$ & Agree\%  \\
\hline
coherence          & 1.080 & 1.270 & +0.190 & 2  & 21 & 77  & 1.000 & 1.060 & +0.060 & 0  & 8  & 92  & 0.090 & 75.0\%  \\
fluency            & 1.460 & 1.960 & +0.500 & 11 & 45 & 44  & 1.224 & 1.690 & +0.466 & 4  & 39 & 57  & 0.280 & 59.0\%  \\
relevance          & 2.250 & 2.140 & -0.110 & 29 & 18 & 53  & 1.571 & 1.630 & +0.059 & 22 & 28 & 50  & 0.195 & 50.0\%  \\
faithfulness       & 1.000 & 1.000 & +0.000 & 0  & 0  & 100 & 1.000 & 1.000 & +0.000 & 0  & 2  & 98  & 0.000 & 98.0\%  \\
informativeness    & 1.020 & 1.190 & +0.170 & 2  & 19 & 79  & 1.000 & 1.010 & +0.010 & 0  & 3  & 97  & 0.035 & 78.0\%  \\
\hline
OVERALL & 1.342 & 1.468 & +0.126 & 26 & 51 & 23  & 1.148 & 1.246 & +0.098 & 21 & 49 & 30  & 0.250 & 53.0\% \\
\hline
\end{tabular}%
}
\caption{Top-$n\sigma$ sampling and + VCM results. Comparison of DeepSeek V3.2 and Qwen3.6-Plus with Cohen's $\kappa$ and agreement statistics. Here, $0$ denotes the baseline sampler and $V$ denotes +VCM.}
\label{tab:topnllmjudge}
\end{table*}

\begin{table*}[!htbp]
\centering
\resizebox{\textwidth}{!}{%
\begin{tabular}{lcccccc|cccccc|cc}
\hline
\multirow{2}{*}{Dimension}
& \multicolumn{6}{c|}{DeepSeek V3.2}
& \multicolumn{6}{c|}{Qwen3.6-Plus}
& \multicolumn{2}{c}{Cohen's $\kappa$ \& Agree.} \\
\cline{2-15}
& Avg(0) & Avg(V) & $\Delta$ & 0 Wins & V Wins & Tie
& Avg(0) & Avg(V) & $\Delta$ & 0 Wins & V Wins & Tie
& $\kappa$ & Agree\%  \\
\hline
coherence          & 1.030 & 1.270 & +0.240 & 0  & 19 & 81  & 1.000 & 1.100 & +0.100 & 0  & 7  & 93  & 0.400 & 86.0\%  \\
fluency            & 1.260 & 1.850 & +0.590 & 10 & 48 & 42  & 1.152 & 1.620 & +0.468 & 4  & 41 & 55  & 0.402 & 66.0\% \\
relevance          & 2.170 & 2.200 & +0.030 & 22 & 26 & 52  & 1.535 & 1.690 & +0.155 & 15 & 29 & 56  & 0.217 & 53.0\%  \\
faithfulness       & 1.000 & 1.010 & +0.010 & 0  & 1  & 99  & 1.000 & 1.000 & +0.000 & 0  & 1  & 99  & -0.010 & 98.0\%  \\
informativeness    & 1.000 & 1.160 & +0.160 & 0  & 12 & 88  & 1.000 & 1.050 & +0.050 & 0  & 6  & 94  & 0.517 & 92.0\%  \\
\hline
OVERALL & 1.281 & 1.462 & +0.181 & 22 & 49 & 29  & 1.130 & 1.264 & +0.134 & 15 & 53 & 32  & 0.398 & 63.0\%  \\
\hline
\end{tabular}%
}
\caption{$p$-less-norm sampling and + VCM results. Comparison of DeepSeek V3.2 and Qwen3.6-Plus with Cohen's $\kappa$ and agreement statistics. Here, $0$ denotes the baseline sampler and $V$ denotes +VCM.}
\label{tab:plessnormllmjudge}
\end{table*}

\begin{table*}[!htbp]
\centering
\resizebox{\textwidth}{!}{%
\begin{tabular}{lcccccc|cccccc|cc}
\hline
\multirow{2}{*}{Dimension}
& \multicolumn{6}{c|}{DeepSeek V3.2}
& \multicolumn{6}{c|}{Qwen3.6-Plus}
& \multicolumn{2}{c}{Cohen's $\kappa$ \& Agree.} \\
\cline{2-15}
& Avg(0) & Avg(V) & $\Delta$ & 0 Wins & V Wins & Tie
& Avg(0) & Avg(V) & $\Delta$ & 0 Wins & V Wins & Tie
& $\kappa$ & Agree\%  \\
\hline
coherence          & 1.060 & 1.230 & +0.170 & 3  & 19 & 78  & 1.010 & 1.071 & +0.061 & 2  & 9  & 89  & 0.132 & 75.0\%  \\
fluency            & 1.350 & 1.850 & +0.500 & 15 & 46 & 39  & 1.224 & 1.485 & +0.261 & 9  & 31 & 60  & 0.279 & 56.0\% \\
relevance          & 2.230 & 2.250 & +0.020 & 21 & 19 & 60  & 1.490 & 1.646 & +0.156 & 13 & 27 & 60  & 0.234 & 57.0\%  \\
faithfulness       & 1.000 & 1.010 & +0.010 & 0  & 1  & 99  & 1.000 & 1.010 & +0.010 & 1  & 3  & 96  & 0.391 & 97.0\% \\
informativeness    & 1.000 & 1.110 & +0.110 & 0  & 9  & 91  & 1.000 & 1.020 & +0.020 & 1  & 4  & 95  & 0.242 & 90.0\%  \\
\hline
OVERALL  & 1.314 & 1.454 & +0.140 & 23 & 50 & 27  & 1.134 & 1.225 & +0.091 & 17 & 46 & 37  & 0.382 & 61.0\%  \\
\hline
\end{tabular}%
}
\caption{Min-$k$ sampling and + VCM results. Comparison of DeepSeek V3.2 and Qwen3.6-Plus with Cohen's $\kappa$ and agreement statistics. Here, $0$ denotes the baseline sampler and $V$ denotes +VCM.}
\label{tab:minkllmjudge}
\end{table*}

\clearpage
\section{Token Rank CDF}
\label{app:tokenrankcdf}

To empirically assess how closely VCM tracks the human token-rank distribution, we plot the Cumulative Distribution Function (CDF) of token ranks. By observing the accumulation of probability mass across token-rank tiers, we can quantitatively measure how a decoding strategy mimics human lexical choices.

\begin{figure*}[!ht]
\centering
\includegraphics[width=\linewidth]{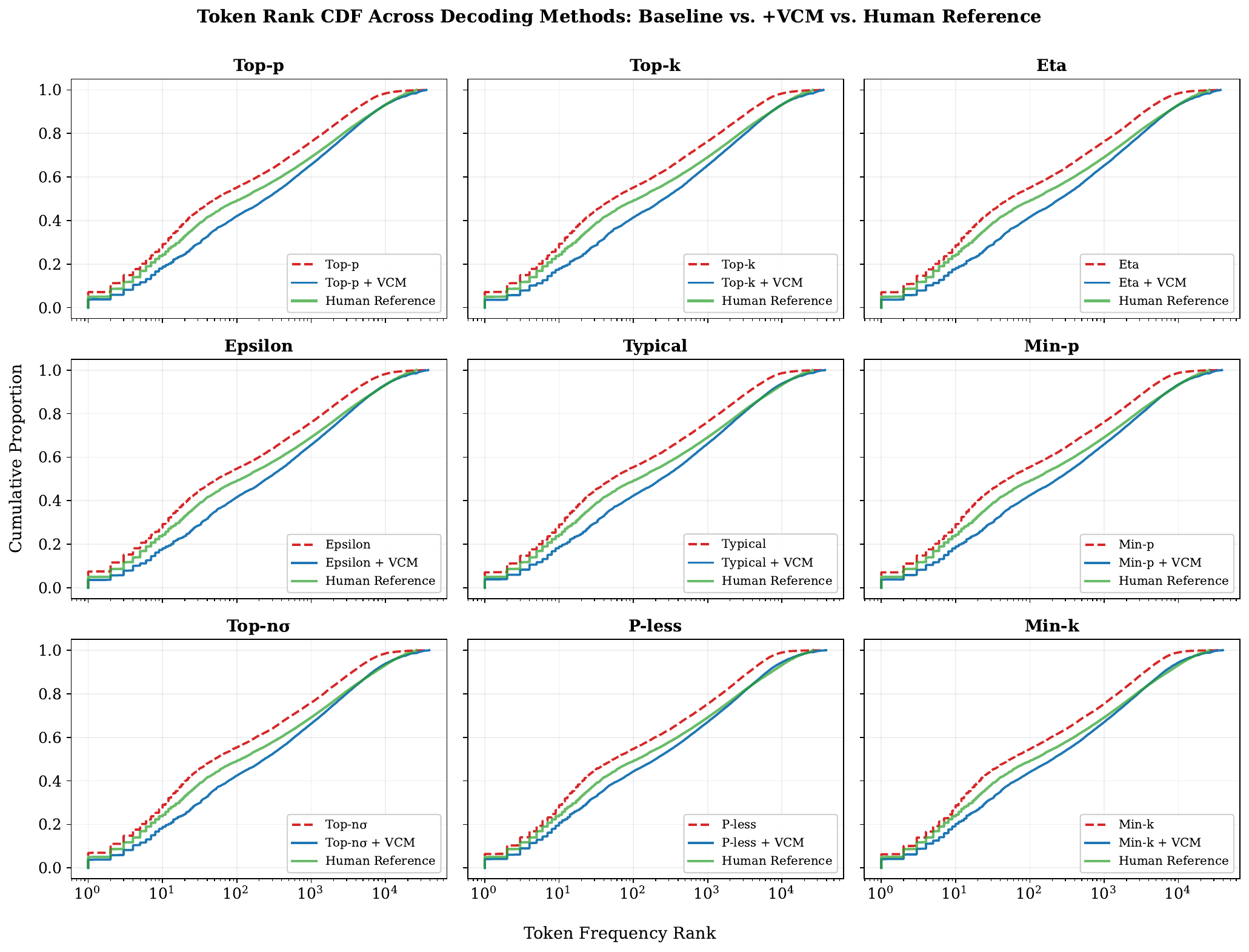}
\caption{Cumulative Distribution Function (CDF) of token ranks across various decoding methods with and without VCM, compared against human reference text. Standard truncation methods exhibit an aggressive, front-heavy bias (steep initial rise), over-relying on top-ranked tokens. In contrast, integrating VCM (solid blue lines) consistently relaxes this strict concentration, reshaping the probability mass to closely track the more distributed, context-aware human reference curve (green lines) across all decoding backbones.
}
\label{fig:tokenrank}
\end{figure*}

To quantify the visual differences, we report three distributional distances. \textbf{KS} is the maximum absolute
vertical gap between two empirical CDFs. \textbf{W1-log} is the
Wasserstein-1 distance computed over $\log_{10}$-transformed token ranks and
measures their overall horizontal displacement. \textbf{CDF Area} is the
normalized absolute area between the two CDFs on the same log-rank axis; it
is therefore a normalized counterpart of W1-log rather than an independent
statistic. Lower values indicate greater similarity to Human Reference. For
each metric, we define
\begin{equation}
    \Delta D = D_{\mathrm{Base}} - D_{\mathrm{+VCM}},
\end{equation}
so a positive $\Delta D$ indicates that VCM reduces the corresponding
distance.

\begin{table*}[t]
    \centering
    \label{tab:distributional_distance}
    \resizebox{\textwidth}{!}{%
    \begin{tabular}{lrrrrrrrrr}
        \toprule
        Method
        & KS Base
        & KS +VCM
        & $\Delta$KS
        & W1-log Base
        & W1-log +VCM
        & $\Delta$W1-log
        & CDF Area Base
        & CDF Area +VCM
        & $\Delta$CDF Area \\
        \midrule
        Top-$p$
        & 0.0720
        & 0.0936
        & -0.0216
        & 0.2325
        & 0.1852
        & 0.0473
        & 0.0506
        & 0.0403
        & 0.0103 \\

        Top-$k$
        & 0.0737
        & 0.1040
        & -0.0302
        & 0.2335
        & 0.2035
        & 0.0300
        & 0.0508
        & 0.0443
        & 0.0065 \\

        $\eta$
        & 0.0735
        & 0.1010
        & -0.0274
        & 0.2306
        & 0.2017
        & 0.0288
        & 0.0502
        & 0.0439
        & 0.0063 \\

        $\epsilon$
        & 0.0713
        & 0.1020
        & -0.0307
        & 0.2288
        & 0.1969
        & 0.0319
        & 0.0498
        & 0.0428
        & 0.0069 \\

        Typical
        & 0.0729
        & 0.0903
        & -0.0174
        & 0.2352
        & 0.1733
        & 0.0619
        & 0.0512
        & 0.0377
        & 0.0135 \\

        Min-p
        & 0.0734
        & 0.0901
        & -0.0167
        & 0.2404
        & 0.1769
        & 0.0635
        & 0.0523
        & 0.0385
        & 0.0138 \\

        Top-n$\sigma$
        & 0.0718
        & 0.0916
        & -0.0198
        & 0.2340
        & 0.1743
        & 0.0598
        & 0.0509
        & 0.0379
        & 0.0130 \\

        $p$-less
        & 0.0746
        & 0.0613
        & 0.0133
        & 0.2169
        & 0.1278
        & 0.0891
        & 0.0472
        & 0.0278
        & 0.0194 \\

        Min-$k$
        & 0.0769
        & 0.0693
        & 0.0076
        & 0.2175
        & 0.1352
        & 0.0823
        & 0.0473
        & 0.0294
        & 0.0179 \\

        \midrule
        \textbf{Mean}
        & \textbf{0.0733}
        & \textbf{0.0892}
        & \textbf{-0.0159}
        & \textbf{0.2299}
        & \textbf{0.1750}
        & \textbf{0.0550}
        & \textbf{0.0500}
        & \textbf{0.0381}
        & \textbf{0.0120} \\
        \bottomrule
    \end{tabular}%
    }
    \caption{Distributional distance to Human Reference.}
\end{table*}

The global metrics show a consistent effect: VCM lowers both W1-log and CDF
Area for all nine decoding methods. Averaged across methods, W1-log decreases
from 0.2299 to 0.1750 and normalized CDF Area decreases from 0.0500 to 0.0381,
corresponding to a 23.9\% reduction in global distributional discrepancy. The
largest reductions occur for P-less and Min-k.

The KS results are more localized. VCM reduces the maximum CDF gap for P-less
and Min-k, but increases it for the other seven methods. This does not
contradict the global improvements: KS depends only on the single largest
vertical deviation, whereas W1-log and CDF Area aggregate differences over the
full rank range. Taken together, the results indicate that VCM consistently
improves the overall token-frequency profile while not uniformly reducing the
worst local mismatch at every rank threshold.

\newpage
\section{Case Study}
\label{app:case}

\begin{table*}[htbp]
\centering

\begin{tabular}{|p{2cm}|p{1.5cm}|p{11.5cm}|}
\hline

\textbf{Question }  & Correct: \textcolor{red}{Athletics}      & New Zealander Jack Lovelock gained fame in the1930's in which field?
 \\
\hline
\textbf{Top-$p$ }   & Predicted: Middle-distance running (1500 meters)    & Let's think step by step! Jack Lovelock was a New Zealander, so we can focus on his achievements in the 1930s. Lovelock was a successful athlete, and he was known for his achievements in a specific sport. He was an Olympic gold medalist and a world champion. After researching, I found that Jack Lovelock won the gold medal in the 1500 meters event at the 1936 Summer Olympics in Berlin, Germany. He was a middle-distance runner and was known for his dominant performance in the 1500 meters. The answer is: \textcolor{red}{The answer is: Middle-distance running (1500 meters)}.
\\
\hline
\textbf{+ CD }      & Predicted: Athletics (Middle-distance running)    & Let's think step by step! Jack Lovelock was a New Zealander, so we can focus on his achievements in the 1930s. Lovelock was a successful athlete, and he was known for his achievements in a specific sport. He was an Olympic gold medalist and a world champion. \textcolor{red}{The answer is: Athletics (Middle-distance running)}

\\
\hline
\textbf{+ DFD  }    & Predicted: Athletics (Track and Field)     & Let's break it down step by step: 1. The question states that Jack Lovelock is a New Zealander. 2. The question mentions that he gained fame in the 1930s. 3. The field in which he gained fame is not specified, but we can try to find more information about him. After researching, I found that Jack Lovelock was a middle-distance runner who won the gold medal in the 1500 meters at the 1936 Summer Olympics in Berlin, Germany. \textcolor{red}{The answer is: Athletics (Track and Field)}
\\
\hline
\textbf{+ VCM   }   & Predicted: \textcolor{red}{Athletics}     & Let me think step by step... Jack Lovelock was a New Zealander... He gained fame in the 1930s... Hmm... He was an athlete of some sort... Wait a minute! I remember now! Jack Lovelock was a middle-distance runner and won a gold medal at the 1936 Summer Olympics in Berlin! \textcolor{red}{The answer is: Athletics}
 \\
\hline
\end{tabular}
\caption{Case Study: Generation examples for the TriviaQA dataset.}
\label{tab:triviaqa case}
\end{table*}

\begin{table*}[t]
\centering

\begin{tabular}{|p{2cm}|p{13cm}|}
\hline

\textbf{Prefix }         & Meridian is home to two post @-@ secondary educational institutions. Meridian Community College, founded in 1937, is located at 910 Highway 19 N and offers, \\
\hline
\textbf{Top-$p$ }        & 910 educational programs. Meridian Community College is located at 910 Highway 19 N and offers 910 educational programs. Meridian Community College is located at 910 Highway 19 N and offers 910 educational programs. Meridian Community College is located at 910 Highway 19 N and offers 910 educational programs. Meridian Community College is located at 910 Highway 19 N and offers 910 educational programs. Meridian Community College is located at 910 Highway 19 N and offers 910 educational programs. Meridian Community College is located at 910 Highway 19 N and offers 910 educational programs. Meridian Community College is located at 910 Highway 19 N and offers 910 educational programs. Meridian Community College is located at 910 Highway 19 N and offers 910 educational programs.\\
\hline
\textbf{+ CD }           & associate degrees. Meridian Community College, located at 910 Highway 19 N, is situated in the state of 910 and provides associate degrees. Meridian Community College, situated at 910 Highway 19 N, is located in the state of 910 and offers associate degrees. Meridian Community College, located at 910 Highway 19 N, is situated in the state of 910 and provides associate degrees. Meridian Community College, located at 910 Highway 19 N, is situated in the state of 910 and offers associate degrees. Meridian Community College, located at 910 Highway 19 N, is situated in the state of 910 and provides associate degrees. Meridian Community College, located at 910 Highway 19 N, is situated in the state of 910 and offers associate degrees.\\
\hline
\textbf{+ DFD  }         & associate degrees. The university is the largest in the state, and the state's largest university is the largest in the state. Meridian is the name of the university, and the university is located at 910 Highway 19 N. The university is the same as the community college, and the community college is the largest in the state. Meridian is a city in the state, and the city is the same as the university. Meridian is the university, and the university is the same as the community college. Meridian is the name of the university, and the university is located at 910 Highway 19 N. Meridian is the largest in the state, and the state's largest university is the largest in the state. Meridian is the name of the university, and the university is located at 910 Highway 19 N. Meridian is the same as the community college, and the community college is the largest in the state. Meridian is a city in the state, and the city is the same as the university. Meridian is the university, and the university is the same as the community college. Meridian is the name of the university, and the university is located at 910 Highway.\\
\hline
\textbf{+ VCM   }        & courses in the sciences and humanities. Meridian Community College has been serving the community since its inception in the year of its founding. The college has a student body that ranges from hundreds to thousands of students each semester. Meridian Community College's campus spans several miles and includes multiple buildings such as laboratories, libraries, and administrative offices. The college's facilities are equipped with modern technology that enhances the learning experience for students from diverse backgrounds and cultures around the world. Meridian Community College's mission is to provide an educational opportunity for all members of the community regardless of their socioeconomic status or other factors that may affect their ability to participate in academic activities. The college's commitment is reflected in its annual report which details the financial performance of the institution over the past fiscal year including revenue generated from tuition fees, grants received from governmental agencies, and expenses incurred in maintaining operational efficiency. \\
\hline
\end{tabular}
\caption{Case Study: Generation examples for the WikiText dataset.}
\label{tab:wikitext case}
\end{table*}

\begin{table*}[htbp]
\centering

\begin{tabular}{|p{2cm}|p{1.5cm}|p{11.5cm}|}
\hline

\textbf{Question }  & Correct \textcolor{red}{2.0}       & John plans to sell all his toys and use the money to buy video games. He has 13 lego sets and he sells them for \$15 each. He ends up buying 8 video games for \$20 each and has \$5 left. How many lego sets does he still have?,
 \\
\hline
\textbf{Top-$p$ }   & Predicted 11.0     & John sells 13 lego sets for \$15 each. He earns 13 * 15 = 195 dollars. He buys 8 video games for \$20 each and has \$5 left, so he spent 20 * 8 + 5 = 165 dollars. Therefore, he spent 195 - 165 = 30 dollars from his lego sets, which means he spent 30 / 15 = 2 lego sets. So, he still has 13 - 2 = \textcolor{red}{The final answer is 11.}
\\
\hline
\textbf{+ CD }      & Predicted 12.0    & John starts with 13 lego sets and sells them for \$15 each. So he gets 13 x 15 = 195 dollars. He buys 8 video games for \$20 each, which costs 8 x 20 = 160 dollars. He has 195 - 160 = 35 dollars left. But he actually has 5 dollars left, not 35. So he must have spent 5 dollars less than he intended. 5 dollars is equal to 5 / 15 = 1/3 of the money he would have gotten from selling one lego set. So he sold 1 lego set.John starts with 13 lego sets and sells 1. So he has 13 - 1 = \textcolor{red}{The final answer is 12.}
\\
\hline
\textbf{+ DFD  }    & Predicted 6.0     & John starts with 13 lego sets. He sells them for \$15 each, so he makes a total of 13 x 15 = 195 dollars. He uses this money to buy 8 video games for \$20 each, which costs 8 x 20 = 160 dollars. He has 195 - 160 = 35 dollars left. He also has 5 dollars left, so he must have spent 35 - 5 = 30 dollars on video games. Since each video game costs 20 dollars, he bought 30 / 20 = 1.5 video games. But he bought 8 video games, so he must have bought 8 - 1.5 = 6.5 video games. He must have sold 6.5 lego sets. He starts with 13 lego sets, so he has 13 - 6.5 = 6.5 lego sets left. \textcolor{red}{The final answer is 6.}
\\
\hline
\textbf{+ VCM   }   & Predicted \textcolor{red}{2.0}     & John sells his lego sets for \$15 each and buys video games for \$20 each, so he makes a profit of \$5 on each set. He ends up with \$5 left over after buying 8 video games at \$20 each. This means he made a total of \$5 + (8 x 20) = \$165 from selling the lego sets. If each set brings in \$15, he sold 165 / 15 = 11 sets. Since he started with 13 sets, he still has: \textcolor{red}{The final answer is 2.}
 \\
\hline
\end{tabular}
\caption{Case Study: Generation examples for the GSM8K dataset.}
\label{tab:gsm8k case}
\end{table*}

\end{document}

%% file: fig_tables/table1_wikitext.tex
\begin{tabular}{lccccccc}
\toprule
\multirow{3}{*}{Method} & \multicolumn{7}{c}{WikiText} \\
\cmidrule(lr){2-8}
& \multirow{2}{*}{Rep-2 $\downarrow$}  & \multirow{2}{*}{Distinct-2 $\uparrow$} & \multirow{2}{*}{MAUVE $\uparrow$}  & \multirow{2}{*}{BERTScore $\uparrow$}  & \multicolumn{3}{c}{LLM-as-a-Judge} \\
\cmidrule(lr){6-8}
& & & & & DeepSeek V3.2 $\uparrow$ & Qwen3.6-Plus $\uparrow$ & Agree. \\
\midrule
Top-$k$                & 79.78$\pm$1.01  & 12.17 & 2.77 & 43.14$\pm$0.30 & 0.24 & 0.26 & \multirow{2}{*}{0.55} \\
\cellcolor{Best}+ VCM   & \cellcolor{Best} \textbf{55.71}$\pm$1.49$^{***}$  & \cellcolor{Best} \textbf{26.75} & \cellcolor{Best} \textbf{16.32} & \cellcolor{Best} \textbf{47.27}$\pm$0.28$^{***}$ & \cellcolor{Best} \textbf{0.59}$^{***}$ & \cellcolor{Best} \textbf{0.53}$^{**}$ &  \\
\midrule
Top-$p$                & 80.59$\pm$1.01  & 17.49 & 2.81 & 42.80$\pm$0.30 & 0.26 & 0.22 & \multirow{2}{*}{0.59} \\
\cellcolor{Best}+ VCM   & \cellcolor{Best} \textbf{58.89}$\pm$1.44$^{***}$  & \cellcolor{Best} \textbf{40.44} & \cellcolor{Best} \textbf{14.71} & \cellcolor{Best} \textbf{46.97}$\pm$0.28$^{***}$ & \cellcolor{Best} \textbf{0.56}$^{**}$ & \cellcolor{Best} \textbf{0.51}$^{***}$ &  \\
\midrule
$\eta$-sampling                & 78.80$\pm$1.08  & 12.37 & 2.46 & 43.25$\pm$0.31 & 0.21 & 0.23 & \multirow{2}{*}{0.53} \\
\cellcolor{Best}+ VCM   & \cellcolor{Best} \textbf{55.22}$\pm$1.48$^{***}$  & \cellcolor{Best} \textbf{26.99} & \cellcolor{Best} \textbf{14.56} & \cellcolor{Best} \textbf{47.43}$\pm$0.29$^{***}$ & \cellcolor{Best} \textbf{0.59}$^{***}$ & \cellcolor{Best} \textbf{0.46}$^{**}$ &  \\
\midrule
$\epsilon$-sampling                & 79.03$\pm$1.08  & 12.34 & 3.47 & 43.19$\pm$0.31 & 0.26 & 0.24 & \multirow{2}{*}{0.62} \\
\cellcolor{Best}+ VCM   & \cellcolor{Best} \textbf{54.83}$\pm$1.49$^{***}$  & \cellcolor{Best} \textbf{27.08} & \cellcolor{Best} \textbf{15.36} & \cellcolor{Best} \textbf{47.41}$\pm$0.28$^{***}$ & \cellcolor{Best} \textbf{0.59}$^{***}$ & \cellcolor{Best} \textbf{0.52}$^{**}$ &  \\
\midrule
Typical                & 81.20$\pm$1.01  & 10.82 & 2.23 & 42.48$\pm$0.29 & 0.19 & 0.16 & \multirow{2}{*}{0.51} \\
\cellcolor{Best}+ VCM   & \cellcolor{Best} \textbf{59.79}$\pm$1.46$^{***}$  & \cellcolor{Best} \textbf{24.69} & \cellcolor{Best} \textbf{14.36} & \cellcolor{Best} \textbf{46.76}$\pm$0.29$^{***}$ & \cellcolor{Best} \textbf{0.62}$^{***}$ & \cellcolor{Best} \textbf{0.55}$^{***}$ &  \\
\midrule
Min-$p$                & 82.70$\pm$0.97  & 10.30 & 1.71 & 41.98$\pm$0.29 & 0.23 & 0.28 & \multirow{2}{*}{0.56} \\
\cellcolor{Best}+ VCM   & \cellcolor{Best} \textbf{61.69}$\pm$1.41$^{***}$  & \cellcolor{Best} \textbf{23.82} & \cellcolor{Best} \textbf{14.40} & \cellcolor{Best} \textbf{46.73}$\pm$0.30$^{***}$ & \cellcolor{Best} \textbf{0.59}$^{***}$ & \cellcolor{Best} \textbf{0.46}$^{*}$ &  \\
\midrule
Top-n$\sigma$                & 82.74$\pm$0.95  & 10.32 & 1.91 & 42.14$\pm$0.29 & 0.26 & 0.21 & \multirow{2}{*}{0.53} \\
\cellcolor{Best}+ VCM   & \cellcolor{Best} \textbf{59.03}$\pm$1.48$^{***}$  & \cellcolor{Best} \textbf{25.13} & \cellcolor{Best} \textbf{13.94} & \cellcolor{Best} \textbf{46.93}$\pm$0.29$^{***}$ & \cellcolor{Best} \textbf{0.51}$^{**}$ & \cellcolor{Best} \textbf{0.49}$^{**}$ &  \\
\midrule
$p\textrm{-less}$-norm                & 85.87$\pm$0.81  & 8.72 & 1.24 & 41.17$\pm$0.30 & 0.22 & 0.15 & \multirow{2}{*}{0.63} \\
\cellcolor{Best}+ VCM   & \cellcolor{Best} \textbf{69.68}$\pm$1.27$^{***}$  & \cellcolor{Best} \textbf{20.01} & \cellcolor{Best} \textbf{10.06} & \cellcolor{Best} \textbf{45.36}$\pm$0.29$^{***}$ & \cellcolor{Best} \textbf{0.49}$^{**}$ & \cellcolor{Best} \textbf{0.53}$^{***}$ &  \\
\midrule
Min-$k$                & 85.98$\pm$0.84  & 8.66 & 1.27 & 41.10$\pm$0.29 & 0.23 & 0.17 & \multirow{2}{*}{0.61} \\
\cellcolor{Best}+ VCM   & \cellcolor{Best} \textbf{75.74}$\pm$1.30$^{***}$  & \cellcolor{Best} \textbf{21.09} & \cellcolor{Best} \textbf{12.68} & \cellcolor{Best} \textbf{45.97}$\pm$0.31$^{***}$ & \cellcolor{Best} \textbf{0.50}$^{**}$ & \cellcolor{Best} \textbf{0.46}$^{***}$ &  \\
\bottomrule
\end{tabular}

%% file: fig_tables/table2_QA.tex
\begin{tabular}{lcccc}
\toprule
\multirow{2}{*}{Method} & \multicolumn{2}{c}{TruthfulQA} & \multicolumn{2}{c}{TriviaQA} \\
\cmidrule(lr){2-3} \cmidrule(lr){4-5}
 & Factuality $\uparrow$ & TTR $\uparrow$ & EM $\uparrow$  & F1 $\uparrow$  \\
\midrule
Top-$k$                & 55.57$\pm$3.41 & 77.59$\pm$0.91 & 49.33$\pm$2.53 & 62.59  \\
\rowcolor{Best}+ VCM   & \textbf{56.79}$\pm$3.40 & \textbf{79.32}$\pm$0.84$^{***}$ & \textbf{52.07}$\pm$2.53 & \textbf{63.29}  \\
\midrule
Top-$p$                & 55.81$\pm$3.41 & 77.04$\pm$0.88 & 49.80$\pm$2.53 & 62.96  \\
\rowcolor{Best}+ VCM   & \textbf{57.16}$\pm$3.39 & \textbf{78.85}$\pm$0.87$^{***}$ & \textbf{52.93}$\pm$2.53 & \textbf{64.63}  \\
\midrule
Typical                & 56.06$\pm$3.40 & 76.88$\pm$0.91 & 48.60$\pm$2.53 & 62.22  \\
\rowcolor{Best}+ VCM   & \textbf{56.43}$\pm$3.40 & \textbf{79.12}$\pm$0.85$^{***}$ & \textbf{51.07}$\pm$2.53 & \textbf{63.28}  \\
\midrule
$\eta$-sampling                & 52.51$\pm$3.42 & 77.54$\pm$0.87 & 47.27$\pm$2.53 & 60.47  \\
\rowcolor{Best}+ VCM   & \textbf{56.30}$\pm$3.40$^{*}$ & \textbf{79.08}$\pm$0.84$^{***}$ & \textbf{51.87}$\pm$2.53$^{*}$ & \textbf{63.54}  \\
\midrule
$\epsilon$-sampling                & 52.88$\pm$3.42 & 77.14$\pm$0.89 & 47.67$\pm$2.53 & 61.36  \\
\rowcolor{Best}+ VCM   & \textbf{55.20}$\pm$3.41 & \textbf{78.95}$\pm$0.84$^{***}$ & \textbf{50.93}$\pm$2.53 & \textbf{63.15}  \\
\midrule
Min-$p$                & 54.59$\pm$3.41 & 77.11$\pm$0.90 & 49.47$\pm$2.53 & 63.18  \\
\rowcolor{Best}+ VCM   & \textbf{56.43}$\pm$3.40 & \textbf{78.80}$\pm$0.84$^{***}$ & \textbf{52.33}$\pm$2.53 & \textbf{63.85}  \\
\midrule
Top-n$\sigma$                & 54.47$\pm$3.41 & 77.06$\pm$0.92 & 48.93$\pm$2.53 & 61.82  \\
\rowcolor{Best}+ VCM   & \textbf{56.79}$\pm$3.40 & \textbf{78.53}$\pm$0.84$^{***}$ & \textbf{51.07}$\pm$2.53 & \textbf{63.02}  \\
\midrule
$p\textrm{-less}$                & 56.30$\pm$3.40 & 76.60$\pm$0.93 & 50.33$\pm$2.53 & 64.61  \\
\rowcolor{Best}+ VCM   & \textbf{57.16}$\pm$3.39 & \textbf{79.35}$\pm$0.86$^{***}$ & \textbf{51.67}$\pm$2.53 & \textbf{64.75}  \\
\midrule
Min-$k$                & 53.98$\pm$3.41 & 76.65$\pm$0.96 & 49.53$\pm$2.53 & 63.10  \\
\rowcolor{Best}+ VCM   & \textbf{56.92}$\pm$3.39 & \textbf{78.34}$\pm$0.90$^{**}$ & \textbf{51.73}$\pm$2.53 & \textbf{63.97}  \\
\bottomrule
\end{tabular}

%% file: fig_tables/table3_reasoning.tex
\begin{tabular}{lcccc}
\toprule
\multirow{2}{*}{Method} & \multicolumn{2}{c}{GSM8K} & \multicolumn{2}{c}{MATH500} \\
\cmidrule(lr){2-3} \cmidrule(lr){4-5}
 & $T=1.0$ & $T=2.0$ & $T=1.0$ & $T=2.0$  \\
\midrule
Top-$k$                & 75.66$\pm$2.32 & 31.69$\pm$2.51 & 18.00$\pm$3.37 & 5.60$\pm$2.02 \\
\rowcolor{Best}+ VCM   & \textbf{77.26}$\pm$2.26 & \textbf{37.53}$\pm$2.61$^{**}$ & \textbf{20.20}$\pm$3.52 & \textbf{6.60}$\pm$2.18 \\
\midrule
Top-$p$                & 73.92$\pm$2.37 & 0.15$\pm$0.21 & 20.40$\pm$3.53 & 0.00\\
\rowcolor{Best}+ VCM   & \textbf{76.65}$\pm$2.28 & \textbf{6.52}$\pm$1.33$^{***}$ & \textbf{22.80}$\pm$3.68 & \textbf{5.20}$\pm$1.95$^{***}$\\
\midrule
Typical                & 74.60$\pm$2.35 & 0.08$\pm$0.15 & 22.00$\pm$3.63 & 0.00 \\
\rowcolor{Best}+ VCM   & \textbf{75.59}$\pm$2.32 & \textbf{6.97}$\pm$1.37$^{***}$ & \textbf{23.80}$\pm$3.73 & \textbf{3.00}$\pm$1.50$^{***}$ \\
\midrule
$\eta$-sampling                & 72.18$\pm$2.42 & 1.90$\pm$0.74 & 18.60$\pm$3.41 & 0.00 \\
\rowcolor{Best}+ VCM   & \textbf{73.69}$\pm$2.38 & \textbf{12.36}$\pm$1.78$^{***}$ & \textbf{21.40}$\pm$3.59 & \textbf{3.80}$\pm$1.68$^{***}$ \\
\midrule
$\epsilon$-sampling                & 72.86$\pm$2.40 & 25.85$\pm$2.36 & 18.60$\pm$3.41 & 5.60$\pm$2.02 \\
\rowcolor{Best}+ VCM   & \textbf{73.01}$\pm$2.40 & \textbf{33.13}$\pm$2.54$^{***}$ & \textbf{23.40}$\pm$3.71 & \textbf{7.20}$\pm$2.27 \\
\midrule
Min-$p$                & 74.14$\pm$2.36 & 62.55$\pm$2.61 & 23.00$\pm$3.69 & 13.80$\pm$3.02 \\
\rowcolor{Best}+ VCM   & \textbf{76.12}$\pm$2.30 & \textbf{65.05}$\pm$2.57 & \textbf{24.60}$\pm$3.78 & \textbf{15.60}$\pm$3.18 \\
\midrule
Top-n$\sigma$                & 76.42$\pm$2.29 & 72.25$\pm$2.42 & 22.40$\pm$3.65 & 19.00$\pm$3.44 \\
\rowcolor{Best}+ VCM   & \textbf{77.10}$\pm$2.27 & \textbf{73.09}$\pm$2.39 & \textbf{24.00}$\pm$3.74 & \textbf{23.60}$\pm$3.72 \\
\midrule
$p\textrm{-less}$                & 77.41$\pm$2.26 & 78.05$\pm$2.23 & 23.20$\pm$3.70 & 23.40$\pm$3.71 \\
\rowcolor{Best}+ VCM   & \textbf{78.54}$\pm$2.22 & \textbf{78.47}$\pm$2.22 & \textbf{25.00}$\pm$3.80 & \textbf{23.80}$\pm$3.73 \\
\midrule
Min-$k$                & 76.27$\pm$2.30 & 75.21$\pm$2.33 & 23.60$\pm$3.72 & 22.00$\pm$3.63 \\
\rowcolor{Best}+ VCM   & \textbf{77.79}$\pm$2.24 & \textbf{76.04}$\pm$2.30 & \textbf{24.20}$\pm$3.75 & \textbf{24.00}$\pm$3.74 \\
\bottomrule
\end{tabular}

%% file: fig_tables/table4_com_DFD.tex
\begin{tabular}{lcccc}
\toprule
\multirow{2}{*}{Method} & \multicolumn{1}{c}{WikiText} & \multicolumn{2}{c}{TriviaQA} & \multicolumn{1}{c}{GSM8K} \\
\cmidrule(lr){2-2} \cmidrule(lr){3-4} \cmidrule(lr){5-5}
 & MAUVE $\uparrow$ & EM $\uparrow$ & F1 $\uparrow$  & EM $\uparrow$  \\
\midrule
Top-$k$                & 2.77 & 49.33$\pm$2.53 & 62.59 & 75.66$\pm$2.32 \\
+ CD                & 3.42 & 51.20$\pm$2.53 & 64.07 & 75.89$\pm$2.31 \\
+ DFD                & 8.32 & 51.40$\pm$2.53 & \textbf{65.00} & 76.95$\pm$2.27 \\
\rowcolor{Best}+ VCM                & \textbf{16.32} & \textbf{52.07}$\pm$2.53 & 63.29 & \textbf{77.26}$\pm$2.26 \\
\midrule
Top-$p$                & 2.81 & 49.80$\pm$2.53 & 62.96 & 73.92$\pm$2.37 \\
+ CD                & 3.41 & 49.93$\pm$2.53 & 63.63 & 75.89$\pm$2.31 \\
+ DFD                & 5.89 & 50.80$\pm$2.53 & 64.42 & \textbf{77.56}$\pm$2.25$^{*}$ \\
\rowcolor{Best}+ VCM                & \textbf{14.71} & \textbf{52.93}$\pm$2.53 & \textbf{64.63} & 76.65$\pm$2.28 \\
\midrule
Min-$p$                & 1.71 & 49.47$\pm$2.53 & 63.18 & 74.14$\pm$2.36 \\
+ CD                & 2.53 & 51.27$\pm$2.53 & 64.05 & 75.81$\pm$2.31 \\
+ DFD                & 4.05 & \textbf{52.60}$\pm$2.53 & \textbf{65.43} & \textbf{77.56}$\pm$2.25$^{*}$ \\
\rowcolor{Best}+ VCM                & \textbf{14.40} & 52.33$\pm$2.53 & 63.85 & 76.12$\pm$2.30 \\
\midrule
$p\textrm{-less}$                & 1.24 & 50.33$\pm$2.53 & 64.61 & 77.41$\pm$2.26 \\
+ CD                & 1.70 & 51.53$\pm$2.53 & 64.51 & 78.32$\pm$2.22 \\
+ DFD                & 2.25 & 51.13$\pm$2.53 & \textbf{65.14} & 78.47$\pm$2.22 \\
\rowcolor{Best}+ VCM                & \textbf{10.06} & \textbf{51.67}$\pm$2.53 & 64.75 & \textbf{78.54}$\pm$2.22 \\
\midrule
Min-$k$                & 1.27 & 49.53$\pm$2.53 & 63.10 & 76.27$\pm$2.30 \\
+ CD                & 1.86 & 49.93$\pm$2.53 & 63.63 & 76.50$\pm$2.29 \\
+ DFD                & 1.38 & 50.67$\pm$2.53 & \textbf{64.57} & \textbf{78.39}$\pm$2.22 \\
\rowcolor{Best}+ VCM                & \textbf{12.68} & \textbf{51.73}$\pm$2.53 & 63.97 & 77.79$\pm$2.24 \\
\bottomrule
\end{tabular}

%% file: fig_tables/table5_speed.tex
\begin{tabular}{lccc}
\toprule
 & \textbf{WikiText} & \textbf{TriviaQA} & \textbf{GSM8K} \\
\midrule
\multicolumn{4}{l}{\textit{Latency (ms/tok) $\downarrow$}} \\[2pt]
Baseline  & 31.33  & 26.55  & 26.84  \\
+ CD      & 48.95  & 37.43  & 41.75  \\
+ DFD      & 85.76  & 67.60  & 67.80  \\
\rowcolor{Best}+ VCM    & 31.88  & 26.86  & 27.18  \\
\midrule
\multicolumn{4}{l}{\textit{Overhead ratio vs.\ baseline}} \\[2pt]
Baseline  & 1.00$\times$  & 1.00$\times$  & 1.00$\times$  \\
+ CD      & 1.62$\times$ & 1.43$\times$  & 1.56$\times$  \\
+ DFD       & 2.74$\times$  & 2.55$\times$  & 2.53$\times$  \\
\rowcolor{Best}+ VCM    & 1.02$\times$  & 1.01$\times$  & 1.01$\times$  \\
\bottomrule
\end{tabular}

%% file: fig_tables/table6_crossmodel.tex
\begin{tabular}{lccc c}
\toprule
& \multicolumn{3}{c}{\textbf{WikiText:}}
& \textbf{GSM8K: Qwen3-} \\
& \multicolumn{3}{c}{\textbf{LLaMA-3-8B}}
& \textbf{4B-Instruct}\\
\cmidrule(lr){2-4}
\cmidrule(lr){5-5}
\textbf{Method}
& \textbf{MAUVE} $\uparrow$
& \textbf{Rep-2} $\downarrow$
& \textbf{BERTScore} $\uparrow$
& \textbf{EM} $\uparrow$ \\
\midrule
Top-$p$
& 84.07
& 16.61
& 57.62
& 92.80 \\
\rowcolor{Best} \quad $+$ VCM
& \textbf{86.62}
& \textbf{12.65}
& \textbf{58.06}
& \textbf{93.33} \\
\midrule
Top-$k$
& 83.00
& 21.52
& 57.37
& 92.80 \\
\rowcolor{Best} \quad $+$ VCM
& \textbf{85.08}
& \textbf{15.52}
& \textbf{58.02}
& \textbf{93.48} \\
\midrule
Min-$p$
& 73.40
& 35.64
& 55.36
& 93.63 \\
\rowcolor{Best} \quad $+$ VCM
& \textbf{87.47}
& \textbf{22.57}
& \textbf{57.53}
& \textbf{93.71} \\
\bottomrule
\end{tabular}